\documentclass[lettersize,journal]{IEEEtran}
\usepackage{amsmath,amsfonts}
\usepackage{algorithmic}
\usepackage{algorithm}
\usepackage{array}
\usepackage[caption=true,font=small]{subfig}
\usepackage{enumitem}
\usepackage{textcomp}
\usepackage{stfloats}
\usepackage{url}
\usepackage{verbatim}
\usepackage{graphicx}
\graphicspath{{./eps}{./png}}
\usepackage{cite}
\usepackage{tikz}
\usetikzlibrary{arrows.meta}
\usepackage{comment}
\usepackage{balance}
\usepackage{hyperref}
\usepackage{algorithmic}
\usepackage{algorithm}

\usepackage{epsfig}
\usepackage{mathrsfs}
\usepackage{mdframed}
\usepackage[acronym]{glossaries}
\usepackage{textcomp}
\usepackage{amssymb}
\usepackage{pifont}

\usepackage{epstopdf}

\begin{document}

\title{\LARGE \bf A Vehicle-in-the-Loop Simulator with AI-Powered Digital Twins for Testing Automated Driving Controllers}

\author{Zengjie Zhang,
        Giannis Badakis$^*$,
        Michalis Galanis$^*$,
        Adem Bavar\c{s}i, Edwin van Hassel,\\
        Mohsen Alirezaei,
        and Sofie Haesaert,~\IEEEmembership{Member,~IEEE}
\thanks{This work is supported by the European project SymAware under grant No. 101070802 and the Dutch NWO Veni project CODEC under grant No. 18244.}
\thanks{Z. Zhang, G. Badakis, M. Galanis, A. Bavar\c{s}i, and S. Haesaert are with the Department of Electrical Engineering, Eindhoven University of Technology,
        PO Box 513, 5600 MB Eindhoven, Netherlands
        (e-mail: z.zhang3@tue.nl, i.badakis@student.tue.nl, m.galanis@student.tue.nl, a.bavarsi@student.tue.nl,
        s.haesaert@tue.nl).
        }
\thanks{E. van Hassel is with Siemens Digital Industries Software, 5708 JZ Helmond, Netherlands (e-mail: edwin.hassel@siemens.com).}
\thanks{M. Alirezaei is with the Department of Mechanical Engineering, Eindhoven University of Technology, PO Box 513, 5600 MB Eindhoven, Netherlands. He is also with Siemens Digital Industries Software, 5708 JZ Helmond, Netherlands (e-mail: m.alirezaei@tue.nl).}     \thanks{
*Equal contributions as secondary authors.
}   
        }



\maketitle

\begin{abstract}
Simulators are useful tools for testing automated driving controllers. Vehicle-in-the-loop (ViL) tests and digital twins (DTs) are widely used simulation technologies to facilitate the smooth deployment of controllers to physical vehicles. However, conventional ViL tests rely on full-size vehicles, requiring large space and high expenses. Also, physical-model-based DT suffers from the reality gap caused by modeling imprecision. This paper develops a comprehensive and practical simulator for testing automated driving controllers enhanced by scaled physical cars and AI-powered DT models. The scaled cars allow for saving space and expenses of simulation tests. The AI-powered DT models ensure superior simulation fidelity. Moreover, the simulator integrates well with off-the-shelf software and control algorithms, making it easy to extend. We use a filtered control benchmark with formal safety guarantees to showcase the capability of the simulator in validating automated driving controllers. Experimental studies are performed to showcase the efficacy of the simulator, implying its great potential in validating control solutions for autonomous vehicles and intelligent traffic. 
\end{abstract}


\section{Introduction}\label{sec:intro}

\IEEEPARstart{I}{n} recent years, advanced automated driving controllers powered by Artificial Intelligence (AI) have facilitated the development of automated driving control algorithms~\cite{yurtsever2020survey}, showing great advantages over conventional heuristic designs due to better learning and reasoning capabilities. For example, deep learning enables the automatic generation of driving controllers from rich interaction data~\cite{wu2021deep}. Logic-based synthesis approaches allow automated reasoning of environmental risks~\cite{lindemann2023risk}. Large language models (LLM) have facilitated the reasoning about human languages~\cite{cui2023drivellm}. However, deploying a trained AI control algorithm to a physical system is challenging due to the \textit{reality gap} which refers to the distinguished characteristics between the physical system and its virtual model used to train the controller~\cite{ngo2021multi}. Thus, simulation is needed to safely validate an AI-powered controller before hardware deployment~\cite{hu2023simulation}, for which a high-fidelity simulation environment with precise virtual models is critical~\cite{zhang2022high, grigorescu2020survey}. 

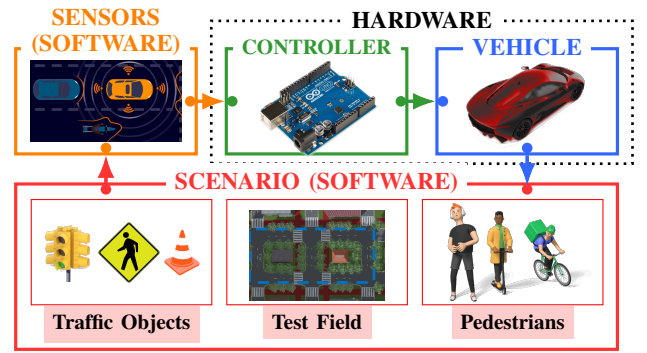
\begin{figure}[htbp]
\noindent
\hspace*{\fill} 
\begin{tikzpicture}[scale=1,font=\small]
\tikzset{arrowdot/.style={-Latex, postaction={draw, circle, fill, minimum size=2cm, inner sep=0pt, at end}}}

\definecolor{darkred}{RGB}{255, 51, 51}
\definecolor{darkblue}{RGB}{51, 101, 255}
\definecolor{darkgreen}{RGB}{51, 153, 51}

\definecolor{shadowgreen}{RGB}{217, 242, 217}
\definecolor{shadowred}{RGB}{255, 204, 204}
\definecolor{shadowblue}{RGB}{204, 221, 255}

\node[minimum height=1.9cm, minimum width=5.6cm, draw=black, dotted, line width=1pt] (bk) at (1.4cm,2.32cm) {};
\node[minimum height=0.1cm, minimum width=0.1cm, anchor=center, align=left, fill=white] at (bk.north) {\textbf{\small HARDWARE}};

\node[minimum height=2.2cm, minimum width=8cm, draw=darkred, line width=1.2pt] (scenario) at (0cm,0cm) {};
\node[minimum height=0.1cm, minimum width=0.1cm, anchor=center, align=left, text=darkred, fill=white] at (scenario.north) {\textbf{\small SCENARIO (SOFTWARE)}};

\node[minimum height=1.4cm, minimum width=2.4cm, anchor=north west,align=left, draw=red] (tl) at ([xshift=-3.8cm, yshift=0.9cm] scenario.center) {};
\node[minimum height=0.1cm, minimum width=0.1cm, anchor=north,align=left, fill=shadowred] at (tl.south) {\textbf{\footnotesize{Traffic Objects}}};
\node[] at ([xshift=-0.7cm] tl.center) {\includegraphics[width=1cm]{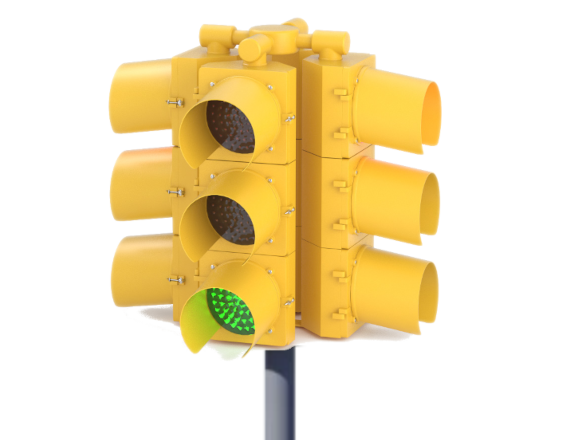}};
\node[] at ([xshift=0.1cm] tl.center) {\includegraphics[width=0.8cm]{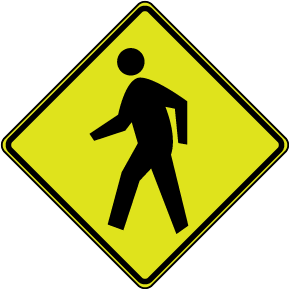}};
\node[] at ([xshift=0.8cm] tl.center) {\includegraphics[width=1cm]{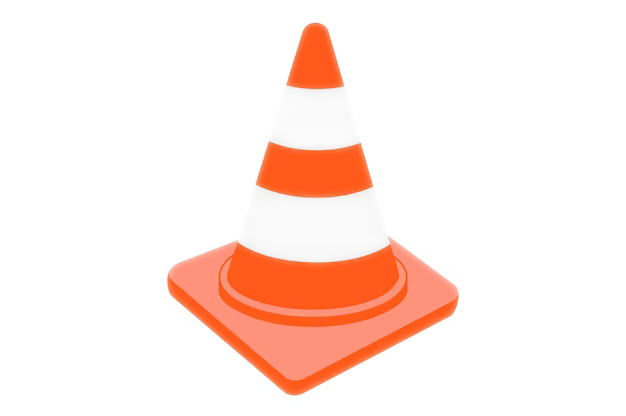}};

\node[minimum height=1.4cm, minimum width=2.4cm, anchor=north west,align=left,draw=red] (tf) at ([xshift=-1.2cm, yshift=0.9cm] scenario.center) {};
\node[minimum height=0.1cm, minimum width=0.1cm, anchor=north, align=left, fill=shadowred] at (tf.south) {\textbf{\footnotesize{Test Field}}};
\node[] at ([yshift=-0.05cm] tf.center) {\includegraphics[width=1.8cm]{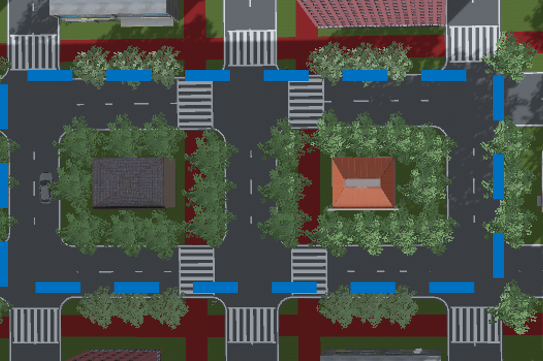}};

\node[minimum height=1.4cm, minimum width=2.4cm, anchor=north west,align=left,draw=red] (pd) at ([xshift=1.4cm, yshift=0.9cm] scenario.center) {};
\node[minimum height=0.1cm, minimum width=0.1cm, anchor=north,align=left, fill=shadowred] at (pd.south) {\textbf{\footnotesize{Pedestrians}}};
\node[] at ([xshift=-0.4cm,yshift=0cm] pd.center) {\includegraphics[width=1.2cm]{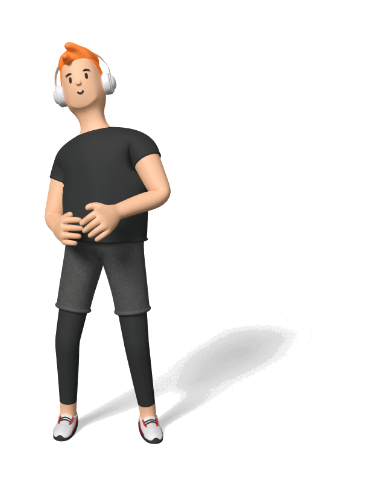}};
\node[] at ([yshift=0.1cm, yshift=-0.1cm] pd.center) {\includegraphics[width=1.2cm]{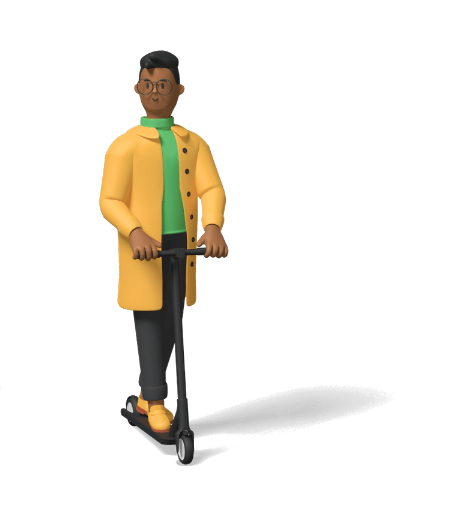}};
\node[] at ([xshift=0.6cm,yshift=0cm] pd.center) {\includegraphics[width=1.2cm]{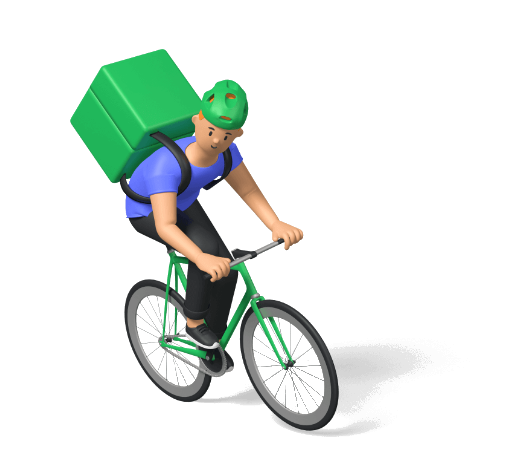}};


\node[minimum height=1.4cm, minimum width=2.4cm, draw=orange, line width=1.2pt] (sensors) at (-2.8cm,2.2cm) {};
\node[minimum height=0.1cm, minimum width=0.1cm, text width=2cm, anchor=center, align=center, fill=white, text=orange] at ([yshift=0.2cm] sensors.north) {\textbf{\small SENSORS \\(SOFTWARE)}};
\node[] at ([yshift=-0cm] sensors.center) {\includegraphics[width=2cm]{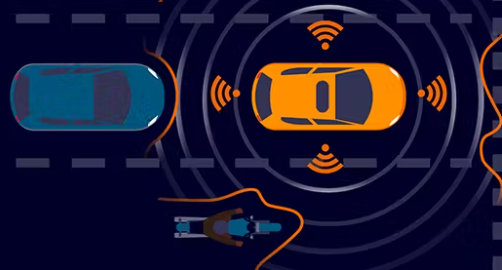}};

\node[minimum height=1.4cm, minimum width=2.4cm, draw=darkgreen, line width=1.2pt] (controller) at (0cm,2.2cm) {};
\node[minimum height=0.1cm, minimum width=0.1cm, anchor=center,align=left, fill=white, text=darkgreen] at (controller.north) {\textbf{\footnotesize CONTROLLER}};
\node[] at ([yshift=-0.1cm] controller.center) {\includegraphics[width=1.6cm]{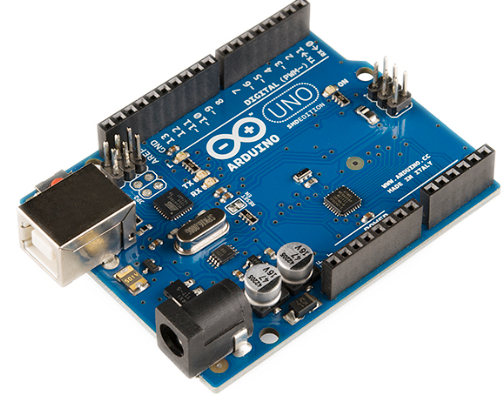}};

\draw[arrowdot, draw=orange, line width=1.2pt] ([xshift=-0.1cm] sensors.east) -- (controller.west);
\fill[fill=orange] ([xshift=-0.1cm] sensors.east) circle (2pt);
\fill[fill=darkgreen] ([xshift=0.1cm] controller.west) circle (2pt);

\node[minimum height=1.4cm, minimum width=2.4cm, draw=darkblue, line width=1.2pt] (vehicle) at (2.8cm,2.2cm) {};
\node[minimum height=0.1cm, minimum width=0.1cm, anchor=center,align=left, text=darkblue, fill=white] at (vehicle.north) {\textbf{VEHICLE}};
\node[] at ([yshift=0cm] vehicle.center) {\includegraphics[width=1.8cm]{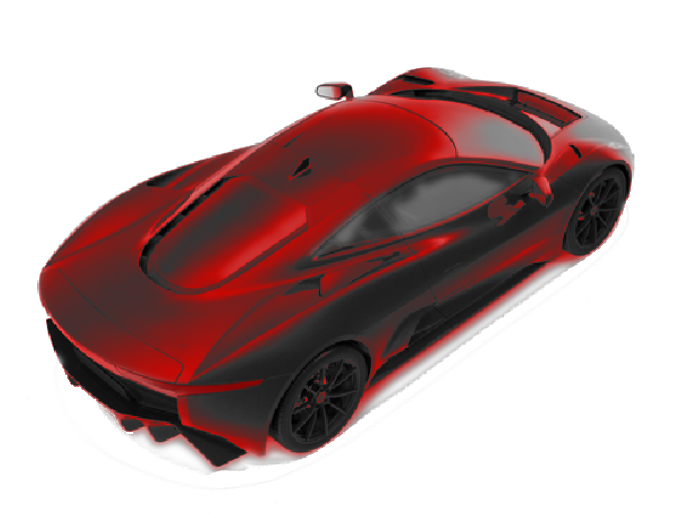}};

\draw[arrowdot, draw=darkgreen, line width=1.2pt] ([xshift=-0.1cm] controller.east) -- (vehicle.west);
\fill[fill=darkgreen] ([xshift=-0.1cm] controller.east) circle (2pt);
\fill[fill=darkblue] ([xshift=0.1cm] vehicle.west) circle (2pt);

\draw[arrowdot, draw=darkred, line width=1.2pt] ([yshift=-0.4cm] sensors.south) -- (sensors.south);
\fill[fill=darkred] ([yshift=-0.45cm] sensors.south) circle (2pt);
\fill[fill=orange] ([yshift=0.1cm] sensors.south) circle (2pt);

\draw[arrowdot, draw=darkblue, line width=1.2pt] (vehicle.south) -- ([yshift=-0.4cm] vehicle.south);
\fill[fill=darkred] ([yshift=-0.45cm] vehicle.south) circle (2pt);
\fill[fill=darkblue] ([yshift=0.1cm] vehicle.south) circle (2pt);

\end{tikzpicture}
\hspace{\fill} 
\caption{The flowchart of vehicle-in-the-loop (ViL) simulation.}
\label{fig:hil}
\end{figure}

Various off-the-shelf simulation software has been provided for testing automated driving controllers. SUMO provides detailed traffic models compatible with real-world data~\cite{kusari2022enhancing}. Open-source simulation engines like CARLA and Gazebo offer sophisticated physical models and high-quality rendering functionalities~\cite{dosovitskiy2017carla, chen2020stabilization}. Commercial simulation software like Siemens Simcenter-Prescan provides integrated development environments with rich sensor libraries and high-fidelity environment models~\cite{ortega2020overtaking}. A survey on simulation software for automated driving can be found in~\cite{li2024choose}. 
However, simulation is not always reliable due to modeling errors. Therefore, hardware-in-the-loop (HIL) tests are performed to smooth the transference of controllers to physical systems~\cite{bullock2004hardware}. HIL refers to substituting the control algorithm in the simulation loop with an onboard control unit using high-speed buses~\cite{landolfi2023hardware}. Similarly, X-in-the-loop~\cite{albers2010implementation} tests refer to substituting a designated simulation component `X' with the corresponding physical entity, such as vehicle-in-the-loop (ViL)~\cite{chen2020mixed}, scenario-in-the-loop~\cite{szalay20205g}, and human-in-the-loop~\cite{zhen2020control}. Among them, ViL, which brings physical vehicles into the simulation loop (Fig.~\ref{fig:hil}), has become an important development phase before road tests~\cite{chen2020mixed}. However, current ViL platforms are mostly built on full-size vehicles, restricting the tests due to space and expenses. A practical and economic simulation platform is needed to validate automated driving controllers efficiently.

Digital twinning (DT) is an emerging technology promising to overcome the limitations of simulation tests of physical vehicles. DT aims to develop a software process that produces similar behaviors to a physical system~\cite{rathore2021role}. Different from physical systems, DT can be easily duplicated, facilitating the extension of simulation to large-scale driving scenarios~\cite{semeraro2021digital, wang2022automatic}. Using DT, reliable results comparable to real hardware tests can be obtained without the limitation of hardware resources. 
In this sense, DT is an ideal technology for testing AI-powered vehicles~\cite{yun2021virtualization, radanliev2022digital}. A perspective review on how DT facilitates AI-powered vehicles can be found in~\cite{wu2022digital}. However, one critical challenge of DT is to ensure the smoothness and reliability of data acquisition, communication, synergy, and storage, such that the virtual processes are sufficiently close to their physical counterparts~\cite{veledar2019digital}. To resolve this issue, learning-based methods have been used~\cite{lv2022artificial}. Nevertheless, the results are difficult to extend and generalize due to overfitting.

The main contribution of this paper is to develop a practical ViL simulator with high-fidelity AI-powered DT models to test automated driving controllers. The ViL mode of the simulator allows users to test benchmarks on physical vehicle hardware in Prescan, a popular and powerful commercial software for simulating autonomous vehicles. The physical vehicle hardware is a 1/10 scaled car that takes up less space and has lower expenses compared to a full-size vehicle. Besides, the DT mode enables users to validate algorithms using high-fidelity vehicle models without hardware. A mixed mode with both hardware and DT allows a flexible extension to large-scale or complex scenarios. Different from the conventional DT, we use a novel data-sequence driven model to facilitate deep learning, ensuring superior modeling precision. The simulator is equipped with two off-the-shelf control benchmarks, namely \textit{Pure Pursuit (PP)} and \textit{Adaptive Cruise Control (ACC)}, allowing users to switch to customized control algorithms. An automated-reasoning-based safety filter is developed to ensure formal guarantees for traffic and safety rules.

The remaining part of the paper is organized as follows. Sec.~\ref{sec:sys} introduces the overall architecture and the detailed components of the simulator. Sec.~\ref{sec:sai} presents the development of the AI-powered DT and the automated-reasoning-based safe controller. In Sec.~\ref{sec:exp}, experimental studies are performed to showcase the efficacy of the simulator. Finally, Sec.~\ref{sec:con} discusses the possible extensions and limitations of the work and concludes the paper.

\textit{Source Code:} the source code of this project is provided in our online GitHub repository~\cite{zhang2024Prescan}.

\section{Simulator Architecture}\label{sec:sys}

This section introduces the architecture of the simulator, followed by a detailed interpretation of its main components. The simulator is deployed on multiple diverse devices connected using a common network. Such a distributed and modularized design allows flexible hardware selection and mode switching according to specific task requirements.

\subsection{Simulator Overview}

\begin{figure}[htbp]
\noindent
\hspace*{\fill} 
\begin{tikzpicture}[scale=1,font=\small]

\definecolor{darkred}{RGB}{255, 51, 51}
\definecolor{darkblue}{RGB}{51, 101, 255}
\definecolor{darkgreen}{RGB}{51, 153, 51}

\definecolor{shadowyellow}{RGB}{255, 255, 229}
\definecolor{shadowgreen}{RGB}{217, 242, 217}
\definecolor{shadowred}{RGB}{255, 204, 204}
\definecolor{shadowblue}{RGB}{204, 221, 255}

\node[minimum height=2.8cm, minimum width=2.2cm, rounded corners=2mm, draw, thick, fill=shadowred] (host) at (-2.2cm,0cm) {};

\node[anchor=north] () at (host.south) {\textbf{Host Computer}};

\node[minimum height=1cm, minimum width=1.8cm, text width=1.2cm, rounded corners=2mm, draw, thick, fill=white, align=center] (Prescan) at ([yshift=0.7cm] host.center) {\textbf{Prescan}};

\node[minimum height=1cm, minimum width=1.8cm, text width=1.2cm, rounded corners=2mm, draw, thick, fill=shadowyellow,align=center] (simulink) at ([yshift=-0.7cm] host.center) {\textbf{ROS\\Interface}};

\draw[<->,>=stealth, dashed, line width=1pt] (Prescan.south) -- node[anchor=west]{API} (simulink.north);


\node[minimum height=1.6cm, minimum width=2.2cm, rounded corners=2mm, draw, thick, fill=shadowgreen] (db) at (-2.2cm,-2.9cm) {\includegraphics[width=1cm]{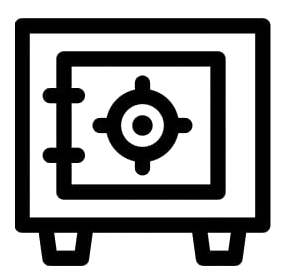}};

\node[anchor=north] () at (db.south) {\textbf{Data Server}};

\draw[->,>=stealth, line width=2pt] ([xshift=0.6cm] db.east) -- (db.east);


\node[minimum height=1.5cm, minimum width=4.3cm, rounded corners=2mm, draw, thick, fill=shadowblue] (target0) at (2.7cm,0.3cm) {};

\node[anchor=south west] () at (target0.north west) {\textbf{ViL Target}};

\node[minimum height=1.0cm, minimum width=1.6cm, text width=1.4cm, align=center, rounded corners=2mm, draw, thick, fill=shadowyellow] (planner0) at ([xshift=-1.15cm, yshift=0cm] target0) {\textbf{Safe\\Controller}};

\node[minimum height=1.0cm, minimum width=1.4cm, text width=1.2cm, rounded corners=2mm, draw, thick, fill=white, align=center] (agent0) at ([xshift=1.2cm, yshift=0cm] target0) {\textbf{Physical\\Vehicle}};

\draw[->, >=stealth, line width=1pt, text width=0.8cm](planner0.east) -- node[pos=0.5, anchor=south, align=center]{$\delta_t$, $u_t$}  (agent0.west);


\node[minimum height=1.5cm, minimum width=4.3cm, rounded corners=2mm, draw, thick, fill=shadowblue] (targetN) at (2.7cm,-2.5cm) {};

\node[anchor=south west] () at (targetN.north west) {\textbf{DT Target}};

\node[minimum height=1.0cm, minimum width=1.6cm, text width=1.4cm, align=center, rounded corners=2mm, draw, thick, fill=shadowyellow] (plannerN) at ([xshift=-1.1cm, yshift=0cm] targetN) {\textbf{Safe\\Controller}};

\node[minimum height=1.0cm, minimum width=1.4cm, text width=1.2cm, rounded corners=2mm, draw, thick, fill=white, align=center] (agentN) at ([xshift=1.2cm, yshift=0cm] targetN) {\textbf{Virtual\\Vehicle}};

\draw[->, >=stealth, line width=1pt, text width=0.8cm](plannerN.east) -- node[pos=0.5, anchor=south, align=center]{$\delta_t$, $u_t$} (agentN.west);


\draw[<->, >=stealth, double, line width=1.5pt]([xshift=-1.3cm, yshift=0.4cm] planner0.west) -- node[pos=0,anchor=south]{ROS}([xshift=-1.3cm, yshift=-1.1cm] plannerN.south west);
\draw[<->,>=stealth, line width=1.5pt] (simulink.east) -- ([xshift=0.7cm] simulink.east);
\draw[<-,>=stealth, line width=1.5pt] (planner0.south) -- ([yshift=-0.5cm] planner0.south);
\draw[->,>=stealth, line width=1.5pt] (agent0.south) -- ([yshift=-0.5cm] agent0.south) -- node[pos=0.78, anchor=north] {\{pose, twist\}}([xshift=-4.5cm,yshift=-0.5cm] agent0.south);
\draw[<-,>=stealth, line width=1.5pt] (plannerN.south) -- ([yshift=-0.5cm] plannerN.south);
\draw[->,>=stealth, line width=1.5pt] (agentN.south) -- ([yshift=-0.5cm] agentN.south) -- node[pos=0.78, anchor=north] {\{pose, twist\}}([xshift=-4.5cm,yshift=-0.5cm] agentN.south);
\draw[->, >=stealth, line width=1.5pt]([xshift=-1.3cm] planner0.west) -- node[pos=0.4, text width=1cm, align=center]{\{sensor, path\}} (planner0.west);
\draw[->, >=stealth, line width=1.5pt]([xshift=-1.3cm] plannerN.west) -- node[pos=0.4, text width=1cm, align=center]{\{sensor, path\}} (plannerN.west);

\end{tikzpicture}
\hspace{\fill} 
\caption{The system architecture of the simulator, where $\delta_t$ and $u_t$ are the command steering angle and velocity of the target vehicle, respectively, and the curly-braced contents \{sensor, path, pose, twist\} refer to the corresponding ROS messages.}
\label{fig:archi}
\end{figure}
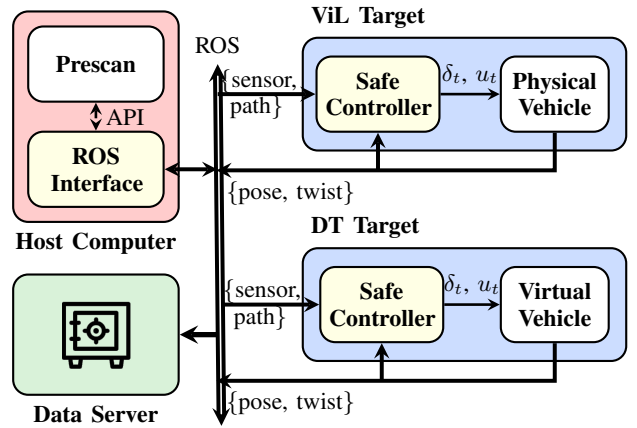

As shown in Fig.~\ref{fig:archi}, the simulator is distributed into multiple devices, including a \textit{Host Computer} running the \textit{Prescan virtual environment}, a \textit{Data Server} storing the simulation data, and multiple \textit{Targets} simulating the behaviors of the \textit{autonomous vehicles}. All devices are connected using a ROS network that transmits the perception information (sensor and path) and the vehicle states (pose and twist) between the host and the targets. The two targets in Fig.~\ref{fig:archi} represent two types of vehicles in the simulation. The \textit{ViL target} denotes a \textit{Physical Vehicle} hardware connected to the network. The \textit{DT target} refers to a \textit{Virtual Vehicle} model programmed on an external computer. Each vehicle is regulated by a steering command $\delta_t$ (in rad) and a velocity command $u_t$ (in m/s) generated by a \textit{Safe Controller} running on the same target device. The subscript $t$ implies that the commands are time-varying. Even though Fig.~\ref{fig:archi} only displays two target devices for brevity, the number of targets can be arbitrarily large if the network allows sufficient bandwidth. The components mentioned above will be interpreted in detail in the successive subsections.

Note that this architecture is consistent with that described in Fig.~\ref{fig:hil}, where the Prescan virtual environment not only renders the driving scenario but also provides virtual sensors for the vehicles. The difference is that the designed simulator is extended with DT targets, allowing virtual vehicles to substitute physical vehicles. This enables a flexible combination of real and virtual vehicles, rendering convenient switching among the following modes, 
\begin{itemize}
    \item \textit{ViL mode:} only physical vehicles are connected;
    \item \textit{DT mode:} only virtual vehicles are connected;
    \item \textit{Mixed mode:} both physical and virtual vehicles are connected.
\end{itemize}
The ViL mode incorporates the physical characteristics of real hardware and is suitable for testing automated driving controllers before hardware deployment. The DT mode provides the highest flexibility by allowing high-fidelity simulation without physical vehicles. The mixed mode enables users to freely increase the number of DT targets for large-scale traffic scenarios with limited hardware resources. The flexible mode switching reflects the versatility and extendability of the simulator to various applications. 

The workflow of the simulator is described as follows.
\begin{itemize}
    \item \textit{Sensor data broadcasting:} The ROS Interface reads the path information and the sensor data of all vehicles from Prescan and publishes them to ROS;
    \item \textit{Vehicle control:} The \textit{Safe Controller} of each target generates the control commands $\delta_t$ and $u_t$ for each vehicle using the sensor data, path information, and the current pose and twist of the vehicle;
    \item \textit{Visualization:} The \textit{ROS Interface} subscribes to the pose and twist of all vehicles and visualizes them in Prescan;
    \item \textit{Data storage:} All ROS messages are recorded in \textit{.bag} files and saved on the Data Server.
\end{itemize}
In this way, both real and virtual vehicles can `see' the scene in the Prescan virtual environment and react accordingly. Their latest poses will be projected in the virtual environment in real time. The users can witness all activities through the Prescan simulation interface. The concurrent mechanism of ROS allows all processes mentioned above to occur asynchronously, which can avoid the influence of network delays. 

\subsection{The Prescan Environment (Host Computer)}\label{sec:Prescan}

The \textit{Host Computer} runs the Prescan virtual environment for the simulator. It is equipped with an Intel(R) Core(TM) i9-10980XE CPU \@ 3.00GHz and an NVIDIA RTX A6000 GPU to ensure decent computation and visualization performance. 
Prescan is a commercial simulation software by Siemens for the development, verification, and validation of automated driving functionalities. It provides a powerful integrated development environment (IDE) that supports a development pipeline from scenario edits to closed-loop tests. The graphic user interface (GUI) gives a convenient and intuitive way to create high-fidelity driving scenarios that resemble the real world. It also has rich libraries for sensors, controllers, traffic objects, and traffic participants with precise descriptions of their real-world characteristics. Moreover, scaling simulation time in Prescan is easy, making it suitable for building ViL and DT simulators. 

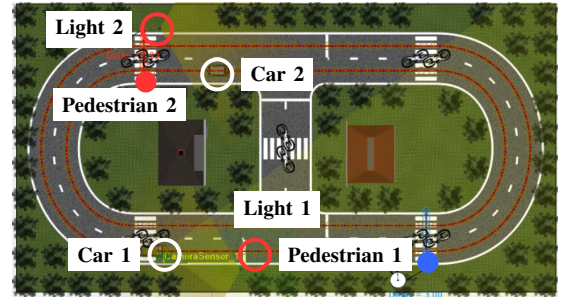
\begin{figure}[htbp]
     \centering

\noindent
\hspace*{\fill} 
\begin{tikzpicture}[scale=1,font=\small]

\definecolor{darkred}{RGB}{255, 51, 51}
\definecolor{darkblue}{RGB}{51, 101, 255}
\definecolor{darkgreen}{RGB}{51, 153, 51}

\definecolor{shadowyellow}{RGB}{255, 255, 229}
\definecolor{shadowgreen}{RGB}{217, 242, 217}
\definecolor{shadowred}{RGB}{255, 204, 204}
\definecolor{shadowblue}{RGB}{204, 221, 255}

\node[] (host) at (-2.2cm,0cm) {\includegraphics[width=0.4\textwidth]{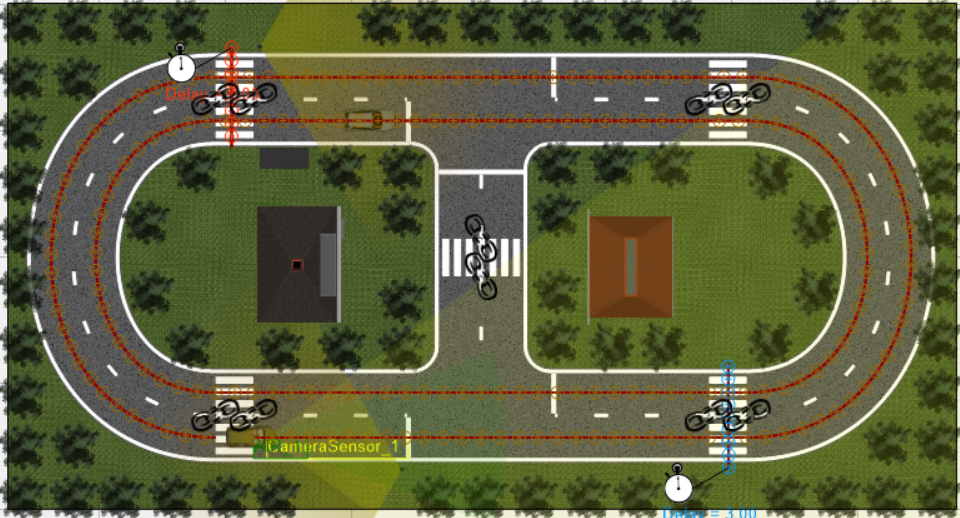}};

\draw[color=darkred, line width=1.5pt] (-3.9cm,1.6cm) circle (0.2cm);
\node[fill=white] (l1) at (-4.8cm,1.6cm) {\textbf{\footnotesize Light 2}};

\draw[color=darkred, line width=1.5pt] (-2.6cm,-1.4cm) circle (0.2cm);
\node[fill=white] (l2) at (-2.3cm,-0.8cm) {\textbf{\footnotesize Light 1}};

\draw[color=white, line width=1.5pt] (-3.1cm,1.0cm) circle (0.2cm);
\node[fill=white] (c1) at (-2.3cm,1.0cm) {\textbf{\footnotesize Car 2}};

\draw[color=white, line width=1.5pt] (-3.8cm,-1.4cm) circle (0.2cm);
\node[fill=white] (c2) at (-4.6cm,-1.4cm) {\textbf{\footnotesize Car 1}};

\node[fill=white] (p1) at (-4.4cm,0.6cm) {\textbf{\footnotesize Pedestrian 2}};
\node[fill=white] (p2) at (-1.4cm,-1.4cm) {\textbf{\footnotesize Pedestrian 1}};
\fill[fill=darkblue] (-0.3cm, -1.5cm) circle (4pt);
\fill[fill=darkred] (-4.05cm, 0.9cm) circle (4pt);

\end{tikzpicture}
\hspace{\fill} 
     
\caption{The benchmarking Prescan scenario.}
     \label{fig:map}
\end{figure}

The basis of the Prescan virtual environment is a map that describes what the vehicles can `see'. Fig.~\ref{fig:map} illustrates the map of the primary benchmarking scenario of our simulator. Sized 70\,m by 30\,m, the scenario simulates a typical neighborhood with an 8-shape bidirectional track. The track has five pedestrian crossings and two T-shape intersections, which are common traffic elements. In the scenario map, traffic lights can be added as traffic objects. Vehicles and pedestrians can be added as primary and secondary traffic participants respectively. Each traffic participant is associated with a path regulating its movement.
Moreover, each vehicle has a front camera to perceive the surrounding environment. Fig.~\ref{fig:map} shows an example with two vehicles, two pedestrians, and two traffic lights. The paths of the vehicles and pedestrians are also highlighted. This simple example can already sufficiently represent a wide range of common practical applications.

\begin{figure}[htbp]
\centering
\subfloat[1st-person perspective.]{\includegraphics[height=2.5cm]{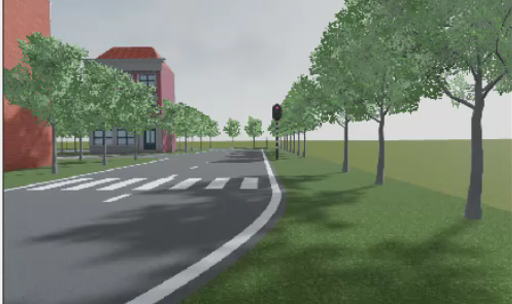}%
\label{fig:1st}}
\hfil
\subfloat[3rd-person perspective.]{\includegraphics[height=2.5cm]{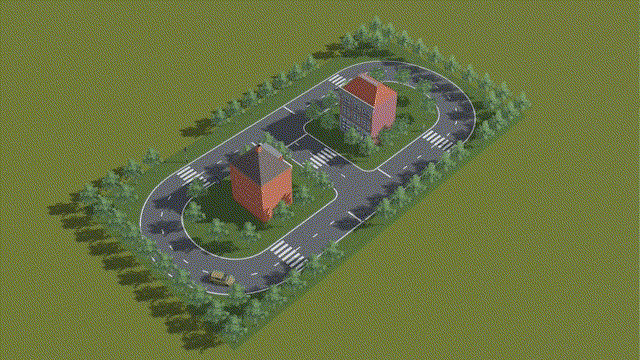}%
\label{fig:3rd}}
\caption{Visualization examples in Prescan.}
\label{fig:views}
\end{figure}

The scenario configurations are saved in a \textit{.pex} which needs to be parsed and built before being used for simulation. The simulation is visualized in a GUI interface which provides various perspectives. Fig.~\ref{fig:views} shows two visualization examples, in the first-person and third-person perspectives, respectively.

\subsection{The ROS Interface (Host Computer)}

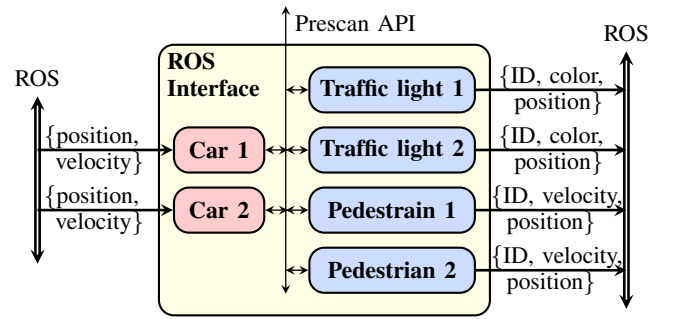
\begin{figure}[htbp]
\noindent
\hspace*{\fill} 
\begin{tikzpicture}[scale=1,font=\small]

\definecolor{darkred}{RGB}{255, 51, 51}
\definecolor{darkblue}{RGB}{51, 101, 255}
\definecolor{darkgreen}{RGB}{51, 153, 51}

\definecolor{shadowyellow}{RGB}{255, 255, 229}
\definecolor{shadowgreen}{RGB}{217, 242, 217}
\definecolor{shadowred}{RGB}{255, 204, 204}
\definecolor{shadowblue}{RGB}{204, 221, 255}

\node[minimum height=3.6cm, minimum width=4.4cm, rounded corners=2mm, draw, thick, fill=shadowyellow] (bridge) at (-2.2cm,0cm) {};
\node[anchor=north west, text width=1.4cm] at (bridge.north west) {\textbf{ROS\\ Interface}};

\node[minimum height=0.6cm, minimum width=1.2cm, rounded corners=2mm, draw, thick, fill=shadowred, align=center] (car1) at ([xshift=-1.4cm, yshift=0.4cm] host.center) {\textbf{Car 1}};
\node[minimum height=0.6cm, minimum width=1.2cm, rounded corners=2mm, draw, thick, fill=shadowred, align=center] (car2) at ([xshift=-1.4cm, yshift=-0.4cm] host.center) {\textbf{Car 2}};

\node[minimum height=0.6cm, minimum width=2.2cm, rounded corners=2mm, draw, thick, fill=shadowblue, align=center] (tl1) at ([xshift=0.9cm, yshift=1.2cm] host.center) {\textbf{Traffic light 1}};
\node[minimum height=0.6cm, minimum width=2.2cm, rounded corners=2mm, draw, thick, fill=shadowblue,align=center] (tl2) at ([xshift=0.9cm, yshift=0.4cm] host.center) {\textbf{Traffic light 2}};

\node[minimum height=0.6cm, minimum width=2.2cm, rounded corners=2mm, draw, thick, fill=shadowblue,align=center] (p1) at ([xshift=0.9cm, yshift=-0.4cm] host.center) {\textbf{Pedestrain 1}};
\node[minimum height=0.6cm, minimum width=2.2cm, rounded corners=2mm, draw, thick, fill=shadowblue, align=center] (p2) at ([xshift=0.9cm, yshift=-1.2cm] host.center) {\textbf{Pedestrian 2}};


\draw[<->,>=stealth, line width=0.5pt] ([xshift=-0.3cm, yshift=0.8cm] tl1.north west) -- node[pos=0, anchor=north west]{Prescan API} ([xshift=-0.3cm, yshift=0cm] p2.south west);

\draw[<->,>=stealth, line width=0.5pt] (car1.east) -- ([xshift=0.3cm] car1.east);
\draw[<->,>=stealth, line width=0.5pt] (car2.east) -- ([xshift=0.3cm] car2.east);

\draw[<->,>=stealth, line width=0.5pt] (tl1.west) -- ([xshift=-0.3cm] tl1.west);
\draw[<->,>=stealth, line width=0.5pt] (tl2.west) -- ([xshift=-0.3cm] tl2.west);
\draw[<->,>=stealth, line width=0.5pt] (p1.west) -- ([xshift=-0.3cm] p1.west);
\draw[<->,>=stealth, line width=0.5pt] (p2.west) -- ([xshift=-0.3cm] p2.west);


\draw[<->,>=stealth, double, line width=1pt] ([xshift=-1.8cm, yshift=0.4cm] car1.north west) -- node[pos=0, anchor=south]{ROS} ([xshift=-1.8cm, yshift=-0.4cm] car2.south west);

\draw[->,>=stealth, line width=1pt] ([xshift=-1.8cm] car1.west) -- node[pos=0.4, text width=1.4cm, align=center]{\{position,\\~~velocity\}} (car1.west);
\draw[->,>=stealth, line width=1pt] ([xshift=-1.8cm] car2.west) -- node[pos=0.4, text width=1.4cm, align=center]{\{position,\\~~velocity\}} (car2.west);


\draw[<->,>=stealth, double, line width=1pt] ([xshift=2cm, yshift=0.2cm] tl1.north east) -- node[pos=0, anchor=south]{ROS} ([xshift=2cm, yshift=-0.2cm] p2.south east);

\draw[->,>=stealth, line width=1pt] (tl1.east) -- node[pos=0.5, text width=1.4cm, anchor=center, align=left]{\{ID, color,\\~~position\}} ([xshift=2cm] tl1.east);
\draw[->,>=stealth, line width=1pt] (tl2.east) -- node[pos=0.5, text width=1.4cm, anchor=center, align=left]{\{ID, color,\\~~position\}} ([xshift=2cm] tl2.east);

\draw[->,>=stealth, line width=1pt] (p1.east) -- node[pos=0.55, text width=1.8cm, anchor=center, align=left]{\{ID, velocity,\\~~position\}} ([xshift=2cm] p1.east);
\draw[->,>=stealth, line width=1pt] (p2.east) -- node[pos=0.55, text width=1.8cm, anchor=center, align=left]{\{ID, velocity,\\~~position\}} ([xshift=2cm] p2.east);

\end{tikzpicture}
\hspace{\fill} 
\caption{The ROS Interface (a Simulink model).}
\label{fig:bridge}
\end{figure}

The \textit{ROS Interface} is a MATLAB Simulink model running on the \textit{Host Computer}, bridging the communication between the Prescan virtual environment and the ROS network. This model is automatically generated from the \textit{.pex} scenario file using Prescan API. Its detailed structure is illustrated in Fig.~\ref{fig:bridge}. Each traffic object or participant defined in the Prescan scenario in Fig.~\ref{fig:map} has a corresponding block in the ROS Interface. For a traffic light, the interface reads its ID, color, and position from the Prescan scenario and publishes the information above to the ROS network. It also publishes the IDs, velocities, and positions of the pedestrians. Meanwhile, the car block subscribes to the position and velocity messages of a vehicle from ROS and sends the information to Prescan. The latter then updates the movement of the vehicle in the virtual environment.

\subsection{The Data Server}

Data play an important role in DT systems since they are fundamental for developing AI-powered approaches. Moreover, the data acquired from hardware devices can help create high-fidelity DT models using data-driven or machine-learning approaches. Data storage is an essential function for the data management of a DT system. Large-scale DT systems usually store data in a cloud server to ensure security~\cite{correia2023data}. The server should be connected to the DT system using a reliable high-bandwidth network to avoid communication losses. The data should be structured in certain formats for the convenience of quick storage and queries, which can be resolved using database technologies~\cite{akhmedova2024structures}. Here, we only prototype a simple data management system fully exploiting the utilities provided by ROS, since a sophisticated database server is unnecessary considering the small amount of simulation data. Specifically, we use a personal laptop as the data server with a ROS node constantly recording the ROS messages into \textit{.bag} files. These data files are regularly uploaded to a cloud server as secure backups. The \textit{rosbag} library allows quick data retrieving using Python. In Sec.~\ref{sec:dnn}, we will demonstrate using the hardware data to train a high-fidelity vehicle DT model with a deep learning method, showcasing a synergy of the physical characteristics between the real and the virtual vehicles.

\subsection{The Scaled Vehicle (ViL Target)}\label{sec:hardware}

\begin{figure}[htbp]
\centering
\subfloat[The F1tenth car]{\includegraphics[height=2.5cm]{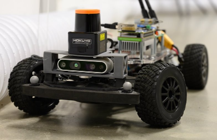}%
\label{fig:car}}
\hfil
\subfloat[The lab space]{\includegraphics[height=2.5cm]{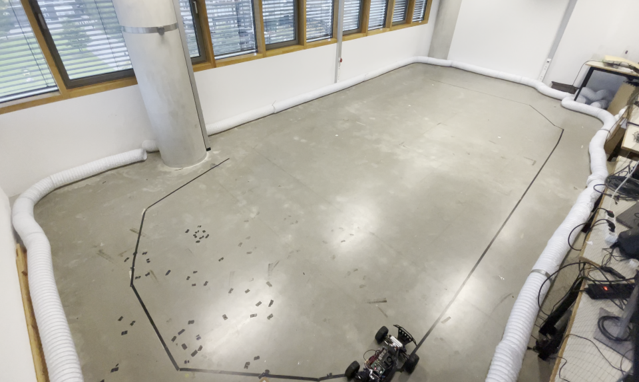}%
\label{fig:lab}}
\caption{The hardware and experimental space.}
\label{fig:hardware}
\end{figure}

Our simulator supports ViL tests by connecting physical vehicles to the ROS network. The benefits of the ViL feature are reflected in two aspects. Firstly, even though simulation studies have become increasingly important for automated driving applications, hardware experiments are still necessary for system validation when the methods are sensitive to the dynamic characteristics of physical vehicles. Secondly, the hardware data in ViL tests can be used to fine-tune the DT models to improve their precision. We use F1tenth scaled cars shown in Fig.~\ref{fig:car} for ViL tests since full-size vehicles are restricted by space and expenses. F1tenth cars are high-performance, 1/10th scale autonomous race cars designed for research and education in robotics and autonomous systems. They offer a hands-on platform for developing, testing, and validating cutting-edge algorithms in perception, planning, and control. An F1tenth car is equipped with advanced sensors and computation units, allowing for real-time processing and decision-making in dynamic environments. We install the Ubuntu 18.04 operating system on the onboard PC of each F1tenth with control algorithms. The size of an F1tenth car fits our 7\,m by 3\,m lab space, which is also 1/10th scale of the Prescan virtual scenario. We use an Epson EB-800F projector to project the top-down view of the scenario onto the ground of the lab, as shown in Fig.~\ref{fig:lab}, rendering an augmented reality (AR) effect. A VICON motion tracking system is used to capture the real-time positions and velocities of the cars and publish them to the ROS network.

\subsection{The Virtual Model (DT Target)}\label{sec:dt}

The virtual vehicle is a modeling program used to simulate the behavior of a scaled vehicle, referred to as its DT. In our case, a virtual vehicle is a ROS node that predicts the next position and velocity of a scaled vehicle given its current state and control command. We compose a simple virtual vehicle model based on the following kinematic model,
\begin{equation}\label{eq:kine}
\left\{\begin{split}
x_{t+1} &= x_{t} + \Delta_t\, v_t \cos(\theta_t) \\
y_{t+1} &= y_{t} + \Delta_t\, v_t \sin(\theta_t) \\
\theta_{t+1} &= \theta_t + \frac{\Delta_t}{l} v_t \tan(\delta_t) \\
v_{t+1} &= v_t + \Delta_t\, a_t, 
\end{split}\right.
\end{equation}
where, $x_t$, $y_t$, $\theta_t$, and $v_t$ are the longitudinal and latitudinal positions, the heading angle, and the longitudinal velocity of the vehicle, respectively, $\delta_t$ is the desired steering command, as introduced before, $a_t$ is the desired acceleration control input calculated using the desired velocity command $u_t$ based on a PD controller $a_t \!=\! K_{\mathrm{p}}(u_t\!-\!v_t) \!+\! K_{\mathrm{d}}(\dot{u}_t\!-\!\dot{v}_t)$, with $K_{\mathrm{p}}$ and $K_{\mathrm{d}}$ being control gains and $\dot{u}_t\!=\!(u_t\!-\!u_{t-1})/\Delta_t$ and $\dot{v}_t\!=\!(v_t\!-\!v_{t-1})/\Delta_t$ as estimated velocities, $l$ is the wheelbase of the car, and $\Delta_t$ is the discrete sampling time. 

Note that the kinematic-based virtual model in Eq.~\ref{eq:kine} is not sufficiently precise for a DT of a scaled vehicle for two reasons. Firstly, the discrete sampling time $\Delta_t$ is difficult to determine due to the non-real-time nature of ROS. The most common way is simply taking the average value of the ROS sampling times, which does not necessarily give us a precise prediction. Secondly, the dynamics of the vehicles are not incorporated in the kinematic model, such as the tire model, the vehicle inertia, frictions, and the dead-zone effects of the mechanical transmission system. 
In Sec.~\ref{sec:dnn}, we will demonstrate a high-fidelity vehicle DT with higher simulation precision using deep learning methods.

\subsection{The Safe Controller (Targets)}\label{sec:safety}

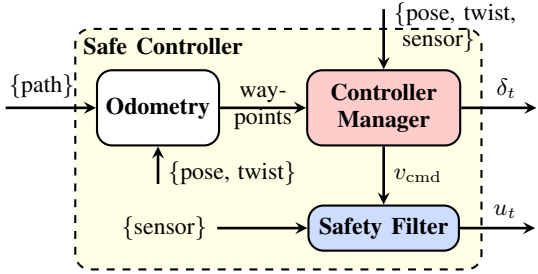
\begin{figure}[htbp]
\noindent
\hspace*{\fill} 
\begin{tikzpicture}[scale=1,font=\small]

\definecolor{darkred}{RGB}{255, 51, 51}
\definecolor{darkblue}{RGB}{51, 101, 255}
\definecolor{darkgreen}{RGB}{51, 153, 51}

\definecolor{shadowyellow}{RGB}{255, 255, 229}
\definecolor{shadowgreen}{RGB}{217, 242, 217}
\definecolor{shadowred}{RGB}{255, 204, 204}
\definecolor{shadowblue}{RGB}{204, 221, 255}

\node[minimum height=3.2cm, minimum width=5.4cm, dashed, rounded corners=2mm, draw, thick, fill=shadowyellow] (bridge) at (-2.2cm,0cm) {};
\node[anchor=north west] at (bridge.north west) {\textbf{Safe Controller}};

\node[minimum height=1.0cm, minimum width=1.6cm, rounded corners=2mm, draw, thick, fill=white, align=center] (odom) at ([xshift=-1.6cm, yshift=0.55cm] host.center) {\textbf{Odometry}};

\node[minimum height=1.0cm, minimum width=2.0cm, text width=1.8cm, rounded corners=2mm, draw, thick, fill=shadowred, align=center] (planner) at ([xshift=1.4cm, yshift=0.55cm] host.center) {\textbf{Controller\\ Manager}};

\draw[->,>=stealth, line width=1pt] (planner.east) -- node[pos=0.6, anchor=south, align=center]{$\delta_t$} ([xshift=1cm] planner.east);

\draw[->,>=stealth, line width=1pt] (odom.east) -- node[pos=0.5, anchor=center, align=center, text width=1cm]{way-\\points} (planner.west);

\node[minimum height=0.6cm, minimum width=2.0cm, rounded corners=2mm, draw, thick, fill=shadowblue,align=center] (safety) at ([xshift=1.4cm, yshift=-1.05cm] host.center) {\textbf{Safety Filter}};

\draw[->,>=stealth, line width=1pt] ([xshift=0cm] planner.south) -- node[pos=0.5, anchor=west, align=right]{$v_{\mathrm{cmd}}$} ([xshift=0cm] safety.north);



\draw[->,>=stealth, line width=1pt] ([xshift=-1.2cm] odom.west) -- node[pos=0.4, anchor=south, align=center]{\{path\}} (odom.west);

\draw[->,>=stealth, line width=1pt] ([yshift=-0.5cm] odom.south) -- node[pos=0.3, anchor=west, text width=1.8cm, align=left]{\{pose, twist\}} (odom.south);


\draw[->,>=stealth, line width=1pt] ([xshift=-1.2cm] safety.west) -- node[pos=0, anchor=east, align=right]{\{sensor\}} (safety.west);

\draw[->,>=stealth, line width=1pt] ([yshift=0.8cm] planner.north) -- node[pos=0.3, anchor=west, align=left] {\{pose, twist, \\~sensor\}} (planner.north);

\draw[->,>=stealth, line width=1pt] (safety.east) -- node[pos=0.6, anchor=south, align=center]{$u_t$} ([xshift=1cm] safety.east);

\end{tikzpicture}
\hspace{\fill} 
\caption{The structure of the \textit{Safe Controller} module, where $\delta_t$ and $v_{\mathrm{cmd}}$ are the command steering angle and velocity of the vehicle, respectively, and $u_t$ is the filtered command velocity.}
\label{fig:controller}
\end{figure}

\begin{figure}[htbp]
\noindent
\hspace*{\fill} 
\begin{tikzpicture}[scale=1,font=\small]

\definecolor{darkred}{RGB}{255, 51, 51}
\definecolor{darkblue}{RGB}{51, 101, 255}
\definecolor{darkgreen}{RGB}{51, 153, 51}

\definecolor{shadowyellow}{RGB}{255, 255, 229}
\definecolor{shadowgreen}{RGB}{217, 242, 217}
\definecolor{shadowred}{RGB}{255, 204, 204}
\definecolor{shadowblue}{RGB}{204, 221, 255}

\node[minimum height=3.4cm, minimum width=5.2cm, dashed, rounded corners=2mm, draw, thick, fill=shadowred] (bridge) at (-0.75cm,-0.3cm) {};
\node[anchor=north west] at (bridge.north west) {\textbf{Controller Manager}};

\node[minimum height=0.6cm, minimum width=1.6cm, draw, thick, fill=white, align=center] (pp) at ([xshift=2cm, yshift=0.6cm] host.center) {\textbf{PP}};

\draw[->,>=stealth, line width=1pt] ([xshift=-2.4cm] pp.west) -- node[pos=0.5, anchor=north, align=center]{\{pose, twist\}} (pp.west);

\draw[->,>=stealth, line width=1pt] ([yshift=0.7cm] pp.north) -- node[pos=0.7, anchor=west, align=center]{way-points} (pp.north);

\draw[<-,>=stealth, line width=1pt] ([xshift=1.4cm] pp.east) -- node[pos=0.45, anchor=south, align=center]{$\delta_t$} (pp.east);


\node[circle,inner sep=0pt,minimum size=0.4cm,draw,thick] (plus) at (-2cm,-0.6cm) {};
\draw[thick] (plus.north east) -- (plus.south west);
\draw[thick] (plus.north west) -- (plus.south east);

\node[minimum height=0.6cm, minimum width=1.6cm, draw, thick, fill=white, align=center] (pid) at ([xshift=2cm, yshift=-0.6cm] host.center) {\textbf{ACC}};

\draw[->,>=stealth, line width=1pt] ([xshift=-1cm] plus.west) -- node[pos=0.4, anchor=south, align=center]{$d_{\min}$} (plus.west);

\draw[->,>=stealth, line width=1pt] (pid.east) -- node[pos=0.45, anchor=south, align=center]{$v_{\mathrm{cmd}}$} ([xshift=1.4cm] pid.east);

\draw[->,>=stealth, line width=1pt] (plus.east) -- (pid.west);

\draw[->,>=stealth, line width=1pt] ([xshift=0.2cm] pid.east) -- ([xshift=0.2cm, yshift=0.6cm] pid.east) -- node[pos=0.7, anchor=center, fill=shadowred, align=center]{$\times t_{\mathrm{safe}}$} ([yshift=0.6cm] plus.center) -- (plus.north);

\draw[->,>=stealth, line width=1pt] ([yshift=-0.3cm] pp.south) -- (pp.south);

\node[circle,inner sep=0pt,minimum size=0.4cm,draw,thick] (plus2) at (0.5cm,-1.4cm) {};
\draw[thick] (plus2.north east) -- (plus2.south west);
\draw[thick] (plus2.north west) -- (plus2.south east);

\draw[->,>=stealth, line width=1pt] ([xshift=1.3cm] plus2.east) -- node[pos=0.65, anchor=south, align=left]{\{sensor\}} (plus2.east);

\draw[->,>=stealth, line width=1pt] ([yshift=-0.6cm] plus2.south) -- node[pos=0.7, anchor=east, align=center]{\{pose\}$-$} (plus2.south);

\draw[->,>=stealth, line width=1pt] (plus2.west) --node[pos=0.5, anchor=south, align=center]{$d_{\mathrm{dist}}$} ([yshift=-0.8cm] plus.center) -- (plus.south);

\node[anchor=south] () at (plus.north east) {};
\node[anchor=north] () at (plus.south east) {\footnotesize\,$-$};
\node[anchor=east] () at (plus.north west) {};

\end{tikzpicture}
\hspace{\fill} 
\caption{The \textit{Controller Manager} equipped with PP and ACC, where $d_{\min}$ and $t_{\mathrm{safe}}$ are the predefined minimal distance and safe time, respectively, and $d_{\mathrm{dist}}$ is the distance between the ego vehicle (provided by \{pose\}) and the obstacle in front (extracted from \{sensor\}). The users are expected to substitute them with customized control algorithms to be tested.}
\label{fig:controller_inside}
\end{figure}
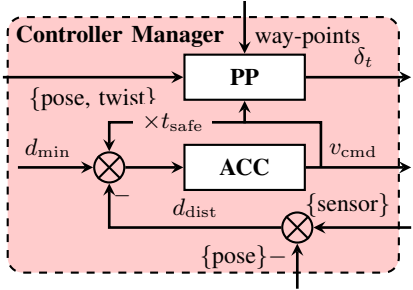

The simulator provides a common \textit{Safe Controller} to generate control commands for the real and virtual vehicles. As illustrated in Fig.~\ref{fig:controller}, The controller consists of a set of ROS nodes including an \textit{Odometry} that updates the waypoints for the vehicle using its up-to-date states, a \textit{Controller Manager} that generates coarse control command $\delta_t$ and $v_{\mathrm{cmd}}$ using the waypoints, and a \textit{safety filter} that filters the velocity command according to the traffic and safety rules. The \textit{Controller Manager} is the core of the \textit{Safe Controller}, maintaining automated driving controllers. The users are allowed to attach customized control algorithms to the \textit{Control Manager} and test them in the simulation. The simulator provides two off-the-shelf control algorithms for path following and speed regulation, respectively, as illustrated in Fig.~\ref{fig:controller_inside}, facilitating the benchmarking of the simulator.

\begin{itemize}
\item \textit{Path following:} for which a Pure Pursuit (PP) controller~\cite{wang2020novel} is used. By selecting a \textit{lookahead point}, a fixed distance ahead of the robot's current position from the desired path, PP adjusts the steering angle $\delta_t$, creating a pursuit-like movement towards this point. PP ensures the robot follows the desired path smoothly by continuously updating the lookahead points. 
\item \textit{Speed regulation:} for which Adaptive Cruise Control (ACC)~\cite{marsden2001towards} is adopted. ACC is an advanced driver assistance system that automatically adjusts a vehicle's speed $v_{\mathrm{cmd}}$ while maintaining a safe distance from any other vehicles driving ahead. It automatically recovers to the normal speed once the road ahead is clear. It has strong adaptability to varying traffic conditions.
\end{itemize}

The control algorithms in the \textit{Control Manager} do not necessarily ensure safety due to the lack of obstacle avoidance. One conventional method for obstacle avoidance is Automatic Emergency Braking (AEB)~\cite{coelingh2010collision}, a safe and automatic braking system to detect and prevent potential crashes, utilizing sensors like radar, lidar, and cameras. It is effective in cases where the driver might not have sufficient reaction times. However, the conventional AEB does not guarantee the compliance of traffic and safety rules, such as \textit{to stop when the light is red or a pedestrian crosses the road}. To resolve this, we design a rule-based safety filter that guarantees the compliance of traffic rules.
With the coarse velocity command $v_{\mathrm{cmd}}$ generated by the Controller, the safety filter generates a filtered velocity command $u_t$ using the following rule-based principle,
\begin{equation}\label{eq:decel}
u_t = \left\{ \begin{array}{ll}
    v_{\mathrm{cmd}} & \mathrm{if}~d > d_{\mathrm{DET}} \\
    v_{\mathrm{dcl}} & \mathrm{if}~d_{\mathrm{EMR}} < d \leq d_{\mathrm{DET}} \\
    0 & \mathrm{if}~d \leq d_{\mathrm{EMR}},
\end{array} \right.
\end{equation}
where $d$ is the distance between the vehicle and a pedestrian or a red traffic light, $d_{\mathrm{DET}}=15\,$m is the minimal distance for the vehicle to react when detecting a pedestrian or red light, $v_{\mathrm{del}}$ is a deceleration velocity to be determined accordingly, and $d_{\mathrm{EMR}}=10\,$m is the maximal emergent distance that the vehicle should perform a full stop. 
This filtering law allows the vehicle to take safe actions in an emergency. Meanwhile, it makes the vehicle decelerate before a full stop in non-emergent cases, ensuring the smoothness of the vehicle's movement.

Such a rule-based filtering law does not scale well for complicated scenarios due to the increasing number of \textit{if} branches. In Sec.~\ref{sec:sfilt}, we will showcase how to automatically compose a safety filter from traffic and safety rules via automated reasoning, without creating \textit{if} rules manually.

\section{Facilitating the Simulator with AI}\label{sec:sai}

The developed simulator has two main issues that may affect its efficacy and scalability. One is the imprecise virtual model which does not fully incorporate the mechanical dynamics of the hardware. The other is the rule-based safety filter which may have scaled up conditional branches as traffic rules increase.
This section introduces improved vehicle DTs and safety filters facilitated by AI-powered approaches.

\subsection{Deep Learning Based Vehicle Model}\label{sec:dnn}

The kinematic-model-based virtual vehicle model introduced in Sec.~\ref{sec:dt} can not precisely characterize the dynamics of a physical vehicle. The non-real-time nature of ROS and the possible network delays also make the discrete sampling time difficult to determine. This may lead to uncertain deviations between the virtual model and the hardware. Out of these concerns, we develop a deep-learning-based DT to improve its simulation precision for a physical vehicle. In this paper, we use a novel data-sequence driven model to facilitate deep learning of the DT.

In a simulation study, we are specifically interested in the motion of a vehicle along its longitudinal velocity $v_t$. Inspecting the virtual model in Eq.~\ref{eq:kine}, the dynamic uncertainties of a scaled vehicle can be incorporated in $a_t$. Thus, we use an unknown function $f(u_t, v_t)$ to substitute $a_t$, leading to
\begin{equation}\label{sec:newkine}
v_{t+1} = v_t + \Delta_t f(u_t, v_t).
\end{equation}
The unknown function $f(u_t, v_t)$ is dependent on $v_t$ since the friction and dead-zone effects are mostly caused by $v_t$. It also depends on $u_t$ to incorporate the effects of the controllers. Although the ground truth of $f$ is difficult to obtain, it is reasonable to assume it as time-invariant and represent it as a parameterized model $f_{\iota}$ with parameter $\iota$.
In the meantime, one can recognize the discrete sampling time $\Delta t$ as a stationary stochastic process, meaning that its expected value $\mathsf{E} (\Delta_t)$ is a time-invariant value $\bar{\Delta}$.

Inspect Eq.~\eqref{sec:newkine} for a continual period of $T$ and consider an initial time $0$ without losing generality. The velocity at time $T$ reads $v_T = v_0 + \sum_{t=0}^{T-1} \Delta_t f_{\iota}(u_t, v_t)$. If the model of $f_{\iota}$ is properly selected, it is possible to determine $\iota_0$, $\iota_1$, $\cdots$, $\iota_{T-1}$, such that $f_{\iota_0}(u_0, v_0) \!=\! \cdots \!=\! f_{\iota_{T-1}}(u_{T-1}, v_{T-1})$, rendering
\begin{equation}\label{eq:newki}
\begin{split}
  \textstyle  v_T &=\textstyle v_0 + \frac{1}{N} (\sum_{t=0}^{T-1} \Delta_t) \sum_{t=0}^{T-1} f_{\iota_t}(u_t, v_t) \\
  & =\textstyle  v_0 + \bar{\Delta} \cdot \mathbf{f}_{\pmb{\iota}}(u_0, v_0, \cdots, u_{T-1}, v_{T-1}),
\end{split}
\end{equation}
where $\bar{\Delta} \!=\! \frac{1}{N} \sum_{t=0}^{T-1} \Delta_t$ is the average sampling interval and $\mathbf{f}_{\pmb{\iota}} = \sum_{t=0}^{T-1} f_{\iota_t}$ can be recognized as a data-sequence driven model of which $\pmb{\iota} = [\,\iota_0~\cdots~\iota_{T-1}\,]^{\top}$ is a parameterization for data sequence $u_0$, $v_0$, $\cdots$, $u_{T-1}$, $v_{T-1}$. The average sampling interval $\bar{\Delta}$ is constant if $\Delta_t$ is stationary and $T$ is sufficiently large. Since $\bar{\Delta}$ is just a constant scalar and $\mathbf{f}_{\iota}$ already depends on $v_0$, we rewrite Eq.~\eqref{eq:newki} as
\begin{equation}
v_T = \mathbf{f}'(\mathbf{u}, \mathbf{v}),
\end{equation}
where $\mathbf{u}:=u_0, u_1, \cdots, u_{T-1}$ and $\mathbf{v}:=v_0,v_1, \cdots, v_{T-1}$ are the history sequences of control command and velocity of the vehicle and $\mathbf{f}' = v_0 + \bar{\Delta} \cdot \mathbf{f}_{\iota}$ is the overall unknown recursive function that derives the velocity $v_T$ using history data $(\mathbf{u}, \mathbf{v})$. 

\begin{figure}[htbp]
     \centering
\includegraphics[width=0.32\textwidth]{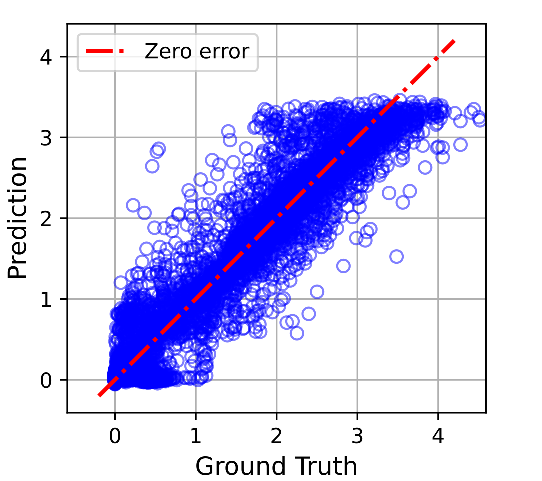}    
\caption{The testing precision of the RNN-based DT.}
     \label{fig:fit}
\end{figure}

The unknown function $\mathbf{f}'$ can be modeled as a recursive neural network (RNN). In the current simulator, we use the following structure that has proved effective for predicting motion sequences of autonomous vehicles~\cite{dang2023incorporating}.
\begin{itemize}
    \item \textit{Input Layer:} a typical input layer with normalized output;
    \item \textit{Encoder Layer:} a gated recurrent unit (GRU) layer with normalized output to encode the dependencies between the successive coordinates of the historical trajectories;
    \item \textit{Latent Feature Layer:} a fully connected layer with Linear rectification functions (ReLU) output for automatic feature extraction;
    \item \textit{Decoder Layer:} a gated recurrent unit (GRU) layer used to decode the sequential features to prediction;
    \item \textit{Output Layer:} a fully connected layer to generate predictions with a 1-dimensional output.
\end{itemize}
We have collected 92077 samples in several hardware experiments to train and validate this model. The hardware experiments have been part of a student course for high-speed navigation challenges. Thus, the vehicle velocities in the data range from 0m/s to 4m/s, sufficiently wide to cover most practical cases. The dataset has been published online at~\cite{zhang2024data}. Note that the model should give zero output when the history data are all zeros. Thus, we augment the samples with another 92077 zero samples to mitigate non-zero predictions with zero inputs. 60\% of the data are used for training, 20\% for an initial validation, and 20\% for the final test. Mean Squared Error (MSE) is used to measure the training cost. The model training program can be found in our online repository~\cite{zhang2024modeling}. The test results of the model are displayed in Fig.~\ref{fig:fit} with the horizontal and vertical representing ground truth and prediction respectively. It shows that the test samples are closely distributed around the zero-error line, implying the high prediction precision of the model.

\subsection{Automated Synthesis of the Safety Filter}\label{sec:sfilt}

The rule-based safety filter introduced in Sec.~\ref{sec:safety} relies on the manual construction of conditional branches, which may scale up for large-scale traffic applications. This subsection proposes an AI-powered method to construct a safety filter automatically from a simple knowledge base. The knowledge base stores the essential traffic rules and safety norms formulated as temporal logic specifications. The safety filter can be automatically synthesized using an off-the-shelf tool.

Let us first elaborate on some essential bases of linear-time temporal logic specifications~\cite{calin2017book} which are important to construct a traffic knowledge base. With $\mathsf{AP}$ denoting a finite set of atomic propositions, an LTL formula is constructed recursively via the syntax $\psi::=\mathrm{true}\,|\,p\,|\,\lnot \psi \, |\, \psi_1 \wedge \psi_2\,|\,\bigcirc \psi\,|\,\psi_1 \mathsf{U} \psi_2$, where $\psi_1$, $\psi_2$, and $\psi$ are LTL formulas, $p\in \mathsf{AP}$ is an atomic proposition, $\lnot$ is the negation operator, $\wedge$ is the conjunction operator, and $\bigcirc$ and $\mathsf{U}$ represent the \textit{next} and \textit{until} temporal operators, respectively. Let $\pmb{\omega}:=\omega_0\omega_1\cdots$ be a string composed of letters from the alphabet $\omega_i \in 2^{\mathsf{AP}}$, for all $i\in\mathbb{N}_{\geq 0}$, with a suffix $\pmb{\omega}_t = \omega_t \omega_{t+1} \omega_{t+2} \dots$, $t \in \mathbb{N}_{\geq 0}$. 
The satisfaction relation denoted as $\pmb{\omega}_t \vDash \psi$ is defined by the following semantics $\pmb{\omega}_t \models p$, if $p \in \omega_t$; $\pmb{\omega}_t \models \lnot p$, if $p \notin \omega_t$;
$\pmb{\omega}_t \models \psi_1 \wedge \psi_2$, if $\pmb \omega_t \models \psi_1$ and $\pmb \omega_t \models \psi_2$;
$\pmb{\omega}_t \models \bigcirc \psi$, if $\pmb\omega_{t+1} \models \psi$;
$\pmb{\omega}_t \models \psi_1 \mathsf{U} \psi_2$, if $\exists$ $i \in \mathbb{N}$ such that $\pmb\omega_{t+i} \models \psi_2$, and $\pmb\omega_{t+j} \models \psi_1$ holds $\forall \, 0\leq j < i$. Based on these essential operators, other logical and temporal operators, namely \textit{disjunction} $\vee$, \textit{implication} $\rightarrow$, \textit{eventually} $\lozenge$, and \textit{always} $\square$ can be defined as, $\psi_1 \vee \psi_2:= \lnot \!\left(\psi_1 \wedge \psi_2 \right)$, $\psi_1 \rightarrow \psi_2$ $:= \lnot \psi_1 \vee \psi_2$, $\lozenge \psi :=\top \mathsf{U} \psi$, and $\square \psi :=\neg \lozenge \neg \psi$.

Incorporating the safety rules elaborated in Sec.~\ref{sec:safety}, we define a set of atomic propositions as $\mathsf{AP}:=\{\mathrm{MOV}$, $\mathrm{DCL}$, $\mathrm{STP}\}$ of which the elements label the driving states of the vehicle, namely \textit{moving in normal speed}, \textit{deceleration}, and \textit{emergency stop}, respectively. We aim to construct a planning-level controller that allows the vehicle to transit between these states subject to the safety rules. Note that a vehicle can not transfer from $\mathrm{STP}$ to $\mathrm{DCL}$. All other state transitions are possible. To address the reaction of the vehicle to the environment, we define another two atomic propositions $\mathrm{URG}$ and $\mathrm{WRN}$ to imply the emergency levels \textit{urgent to stop} and \textit{warning to decelerate}, where $\mathrm{URG}$, $\mathrm{WRN}$ and $\lnot(\mathrm{URG} \!\vee\! \mathrm{WRN})$ correspond to $d \leq d_{\mathrm{EMR}}$, $d > d_{\mathrm{DET}}$, and $d_{\mathrm{EMR}} < d \leq d_{\mathrm{DET}}$, respectively. Then, the traffic safety rules consistent with the rule-based law in Sec.~\ref{eq:decel} can be represented as the following GR(1) form LTL formula $\varphi$,
\begin{equation*}
\begin{split}
\varphi:= &\, \varphi_e \rightarrow \varphi_v \\
    \varphi_e:= & \, (\square \lozenge \lnot \mathrm{URG}) \wedge (\square \lozenge \lnot \mathrm{WRN}) \\
    \varphi_v :=&\, \square \lozenge \mathrm{MOV} \wedge \square ((\mathrm{URG} \wedge \lnot \mathrm{WRN}) \rightarrow \bigcirc \mathrm{STP}) \\
    & \wedge \square ((\mathrm{WRN} \wedge \lnot \mathrm{URG} \wedge (\mathrm{MOV} \vee \mathrm{DCL})) \rightarrow \bigcirc \mathrm{DCL}) \\
    & \wedge \square (\lnot (\mathrm{URG} \vee \mathrm{WRN}) \rightarrow \bigcirc \mathrm{MOV}),
\end{split}
\end{equation*}
where \textit{GR} refers to \textit{Generative Reaction} and $\psi_e$ and $\psi_v$ represent the environment's and the vehicle's specifications, respectively. The formula implies an \textit{assume-guarantee} contract, meaning that the vehicle's behavior should satisfy $\psi_v$ given that the environment satisfies $\psi_e$. Both $\psi_e$ and $\psi_v$ are in a Conjunctive Normal Form (CNF) in which each subformula serves as a knowledge item stored in the knowledge base. Thus, users only need to add new knowledge items when the traffic scenario becomes more complicated, instead of manually rewriting the rule-based conditional structure, improving the scalability and flexibility of the controller design. A high-level controller regulating the desired vehicle state transition can be automatically synthesized using the \textit{TuLip} toolbox~\cite{filippidis2016control}. The source code to solve this safety filter can be found in our GitHub reponsitory~\cite{zhang2024sm}.

\begin{figure}[htbp]
     \centering
\includegraphics[width=0.48\textwidth]{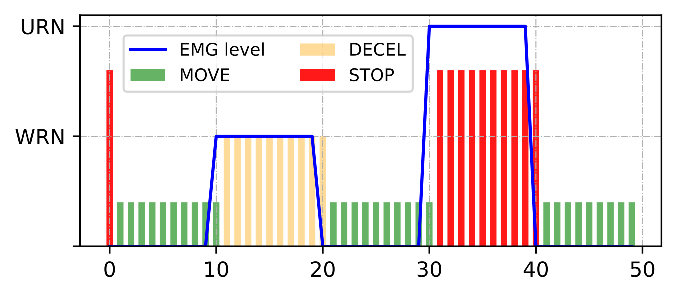}    
\caption{The temporal sequence of vehicle states.}
     \label{fig:seq}
\end{figure}

Fig.~\ref{fig:seq} shows the efficacy of the synthesized safety filter in a simple simulation study. We consider a test with 50 steps, where the vehicle starts from a $\mathrm{STP}$ state, sees a pedestrian crossing the road at a distance $d_{\mathrm{EMR}} < d \leq d_{\mathrm{DET}}$ ($\mathrm{WRN}$) from step 10 to step 20, and is confronted with a red light at a distance $d < d_{\mathrm{EMR}}$ ($\mathrm{URN}$) between steps 30 and 40. It can be seen that the vehicle decelerates when $\mathrm{WRN}$ is true and performs a full stop when $\mathrm{URG}$ becomes true. Once $\mathrm{WRN}$ or $\mathrm{URG}$ is released, the vehicle immediately recovers to normal movement. This indicates that the automatedly generated safety filter can ensure safe driving behavior.

In this example, constructing the LTL specifications seems slightly more complicated than the rule-based law in Eq.~\ref{eq:decel}. However, the superior efficiency of automated synthesis will be significant for large-scale traffic scenarios~\cite{maierhofer2022formalization}.

\section{Experimental Validation}\label{sec:exp}

This section uses two experiments to evaluate the efficacy of the designed simulator. The first one verifies that the behavior of the DT model is consistent with a real scaled car, and the second one uses a demonstration to showcase how the simulator can be potentially extended to complex simulation scenarios utilizing its ViL and DT features. Both experiments are conducted in a Prescan\textregistered~scenario illustrated in Fig.~\ref{fig:map}. The video demonstrating the experiments can be found in \href{https://youtu.be/aTvu2ilaggw}{https://youtu.be/aTvu2ilaggw}.

\subsection{Validation of DT models}\label{sec:val_dt}

The first experiment compares a kinematic-based model, an RNN-based model, and a real scaled vehicle of Car 1, and set a dummy virtual model as Car 2. Light 1 is a regular traffic light and Light 2 is used to remind the vehicles about any crossing pedestrians. Both lights are initially set to red, and turn to yellow and green after a time duration of 5 second and 2 second, respectively. Both cars are required to drive along the round tracks of the map, with Car 1 moving anti-clockwise along the outer track and Car 2 moving clockwise along the inner track. Both cars are equipped with the safe controllers described in Sec.~\ref{sec:sfilt}, ensuring them to stop when seeing a red light and resume moving when the light turns green. Two pedestrians are scheduled to cross the road along the pedestrian crossings when the cars are approaching, in a way to test their capabilities of pedestrian avoidance.

\begin{figure}[htbp]
\centering
\subfloat[]{
\begin{tikzpicture}[scale=1,font=\small]%
\node[] (host) at (0cm,0cm) {\includegraphics[height=2.3cm]{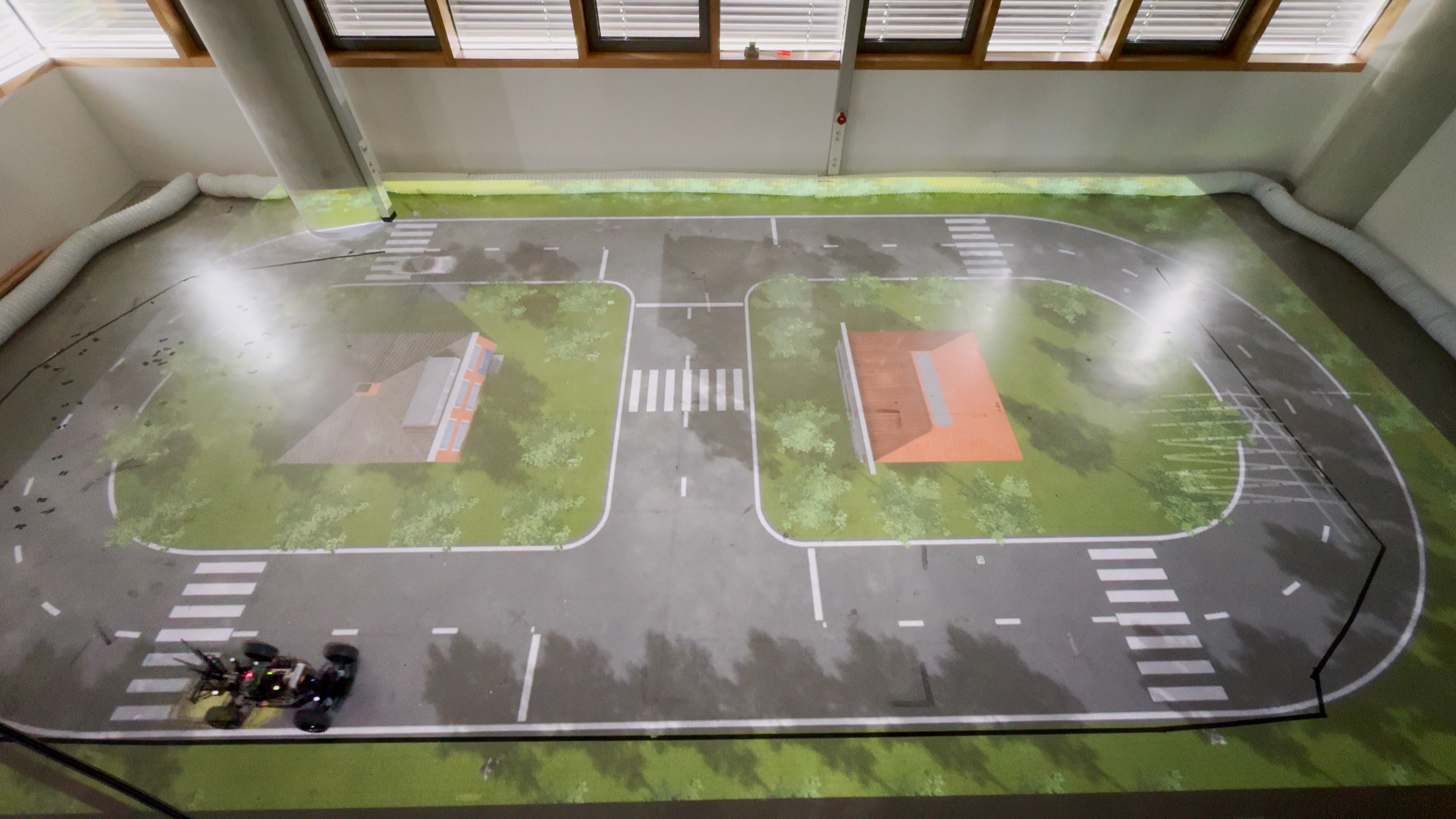}};
\node[minimum height=0.4cm, minimum width=0.4cm, draw, thick, blue] () at (-1.3cm,-0.8cm) {};
\node[minimum height=0.4cm, minimum width=0.4cm, draw, thick, red] () at (-0.8cm,0.4cm) {};
\node[] (host) at (-0.7cm,0.9cm) {\includegraphics[width=0.2cm]{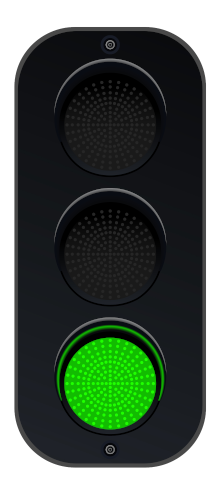}};
\node[] (host) at (-0.5cm,-0.7cm) {\includegraphics[width=0.2cm]{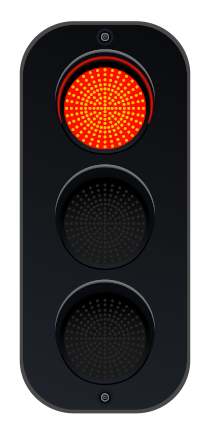}};
\end{tikzpicture}
\label{fig:1a}
}
\hspace{-0.4cm}
\subfloat[]{
\begin{tikzpicture}[scale=1,font=\small]%
\node[] (host) at (0cm,0cm) {\includegraphics[height=2.3cm]{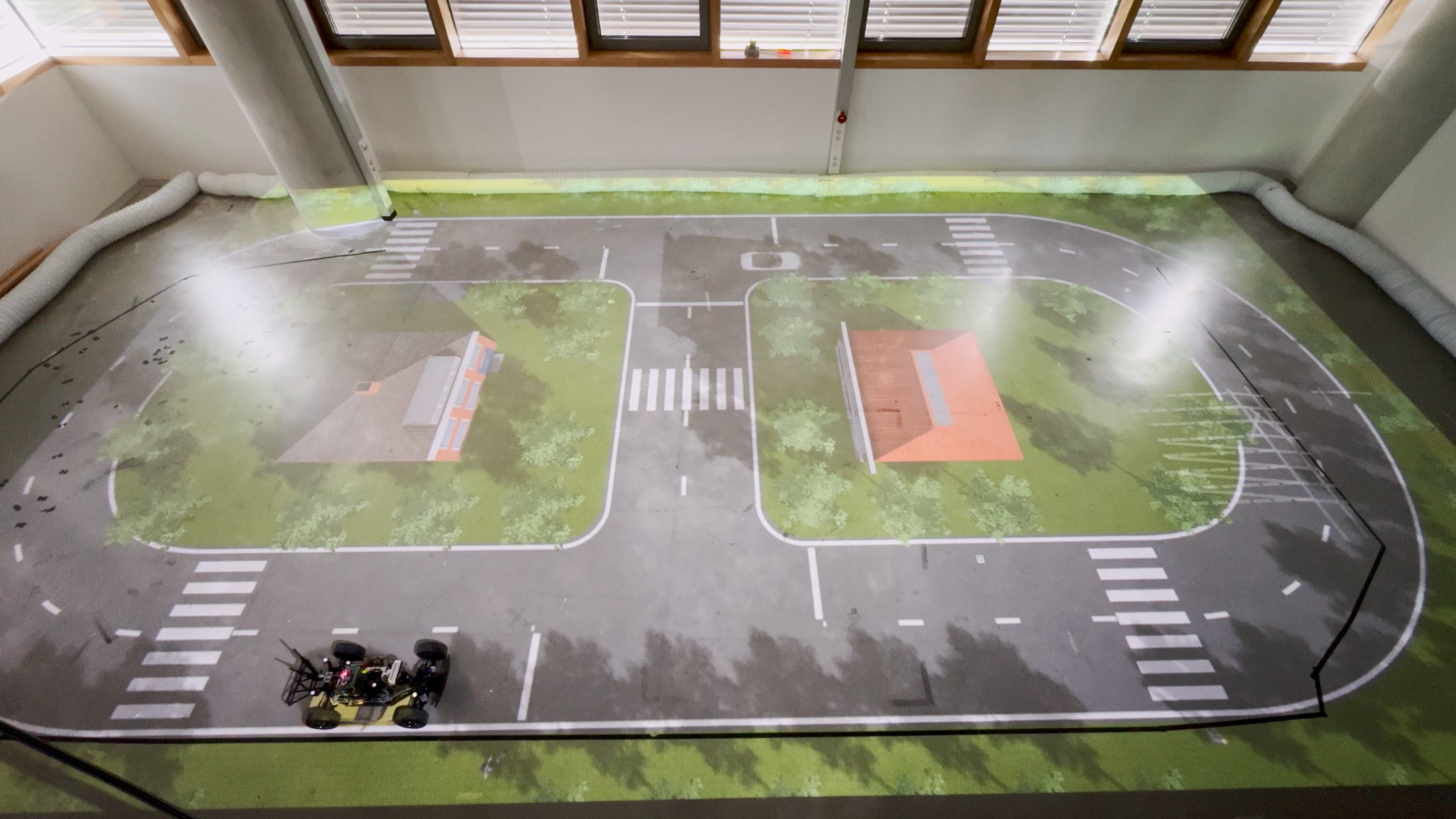}};
\node[minimum height=0.4cm, minimum width=0.4cm, draw, thick, blue] () at (-1cm,-0.8cm) {};
\node[minimum height=0.4cm, minimum width=0.4cm, draw, thick, red] () at (0.1cm,0.4cm) {};
\node[] (host) at (-0.7cm,0.9cm) {\includegraphics[width=0.2cm]{green.png}};
\node[] (host) at (-0.5cm,-0.7cm) {\includegraphics[width=0.2cm]{red.png}};
\end{tikzpicture}
\label{fig:1b}
}
\hspace{-0.4cm}
\subfloat[]{
\begin{tikzpicture}[scale=1,font=\small]%
\node[] (host) at (0cm,0cm) {\includegraphics[height=2.3cm]{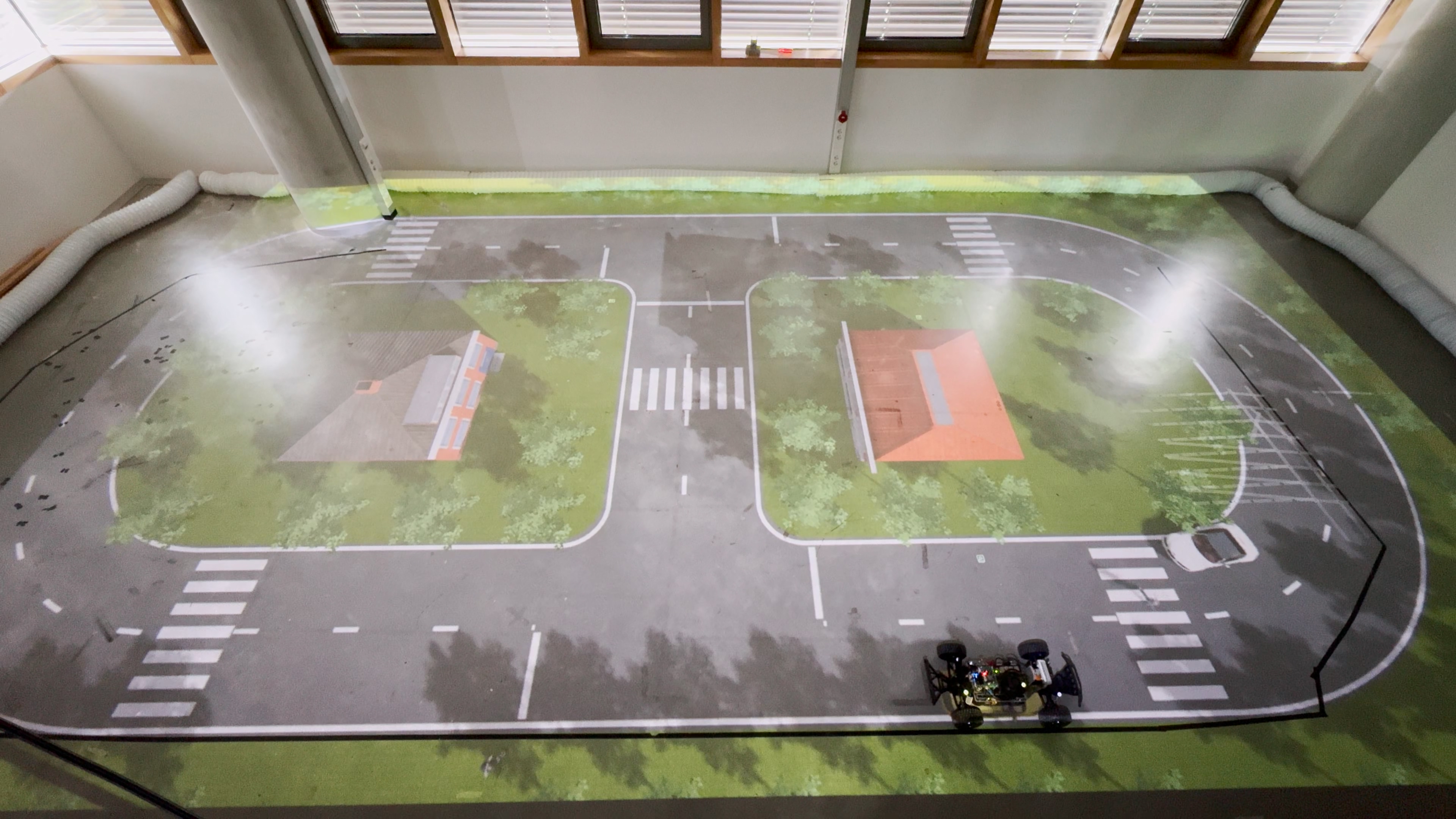}};
\node[minimum height=0.4cm, minimum width=0.4cm, draw, thick, blue] () at (0.8cm,-0.8cm) {};
\node[minimum height=0.4cm, minimum width=0.4cm, draw, thick, red] () at (1.4cm,-0.4cm) {};
\node[] (host) at (-0.7cm,0.9cm) {\includegraphics[width=0.2cm]{red.png}};
\node[] (host) at (-0.5cm,-0.65cm) {\includegraphics[width=0.2cm]{green.png}};
\end{tikzpicture}
\label{fig:1c}}
\hspace{-0.4cm}
\subfloat[]{
\begin{tikzpicture}[scale=1,font=\small]%
\node[] (host) at (0cm,0cm) {\includegraphics[height=2.3cm]{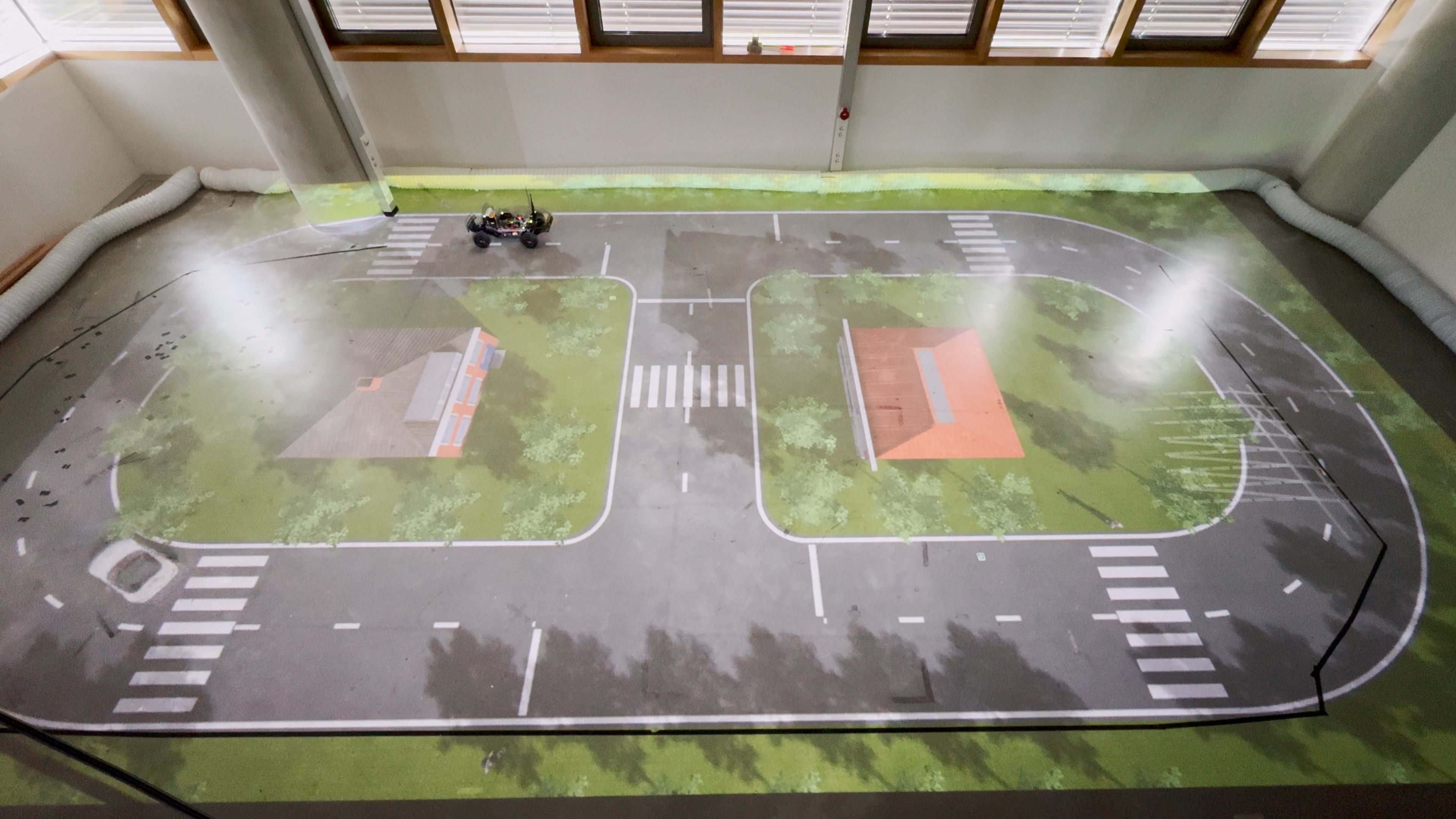}};
\node[minimum height=0.4cm, minimum width=0.4cm, draw, thick, blue] () at (-0.6cm,0.5cm) {};
\node[minimum height=0.4cm, minimum width=0.4cm, draw, thick, red] () at (-1.6cm,-0.5cm) {};
\node[] (host) at (-0.7cm,0.9cm) {\includegraphics[width=0.2cm]{red.png}};
\node[] (host) at (-0.5cm,-0.65cm) {\includegraphics[width=0.2cm]{green.png}};
\end{tikzpicture}
\label{fig:1d}}
\caption{Experiment to test the DT model and the safe controller in a scenario with two vehicles. (a) The initial configuration. (b) Car 2 (embraced by a blue frame) stops when Light 2 is red and resumes moving when the light becomes green. (c) Car 1 (embraced by a red frame) and Car 2 stop for a pedestrian crossing the road. (d) Car 2 stops when Light 1 becomes red.}
\label{fig:views1}
\end{figure}

\begin{figure}[htbp]
     \centering
\includegraphics[width=0.48\textwidth]{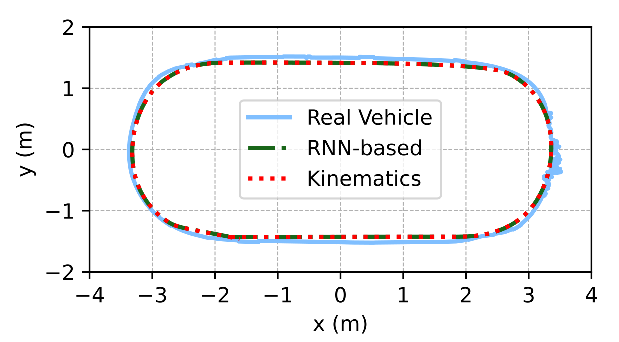}    
\caption{The comparison between the trajectories of the scaled vehicle, the RNN-based virtual model, and the kinematic-based virtual model.}
     \label{fig:traj}
\end{figure}

\begin{figure}[htbp]
     \centering
\includegraphics[width=0.47\textwidth]{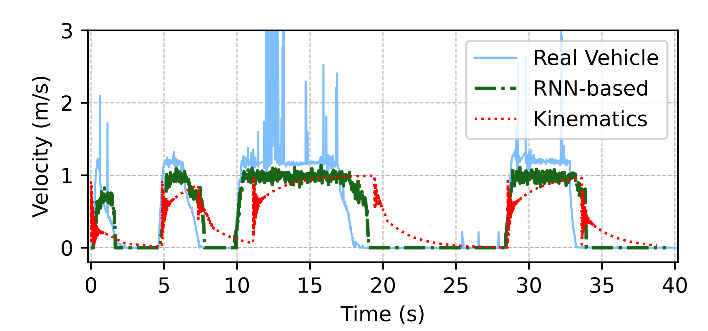}    
\caption{The comparison between the velocities of the scaled vehicle, the RNN-based virtual model, and the kinematic-based virtual model.}
     \label{fig:vel}
\end{figure}

Fig.~\ref{fig:views1} shows that both cars successfully achieve the driving task while obeying the traffic rules and yielding to the pedestrians. This implies the efficacy of the designed safe controller. Fig.~\ref{fig:traj} displays the trajectories of the scaled vehicle hardware, the RNN-based model, and the kinematic-based model of Car 2. It shows that the three trajectories coincide with each other, indicating the precision of the RNN-based and the kinematic-based virtual models. Fig.~\ref{fig:vel} gives a deeper insight to the precision of these two virtual models by showing their velocities. It can be seen that the RNN-based model fits the scaled vehicle hardware better than the kinematic-based model. This implies that the RNN-based model is more precise than the kinematic-based model, making it more suitable as a DT of the scaled vehicle.

\subsection{Validation of Safety Filter}\label{sec:valdt}

\begin{figure}[htbp]
\centering
\subfloat[]{
\begin{tikzpicture}[scale=1,font=\small]%
\node[] (host) at (0cm,0cm) {\includegraphics[height=2.3cm]{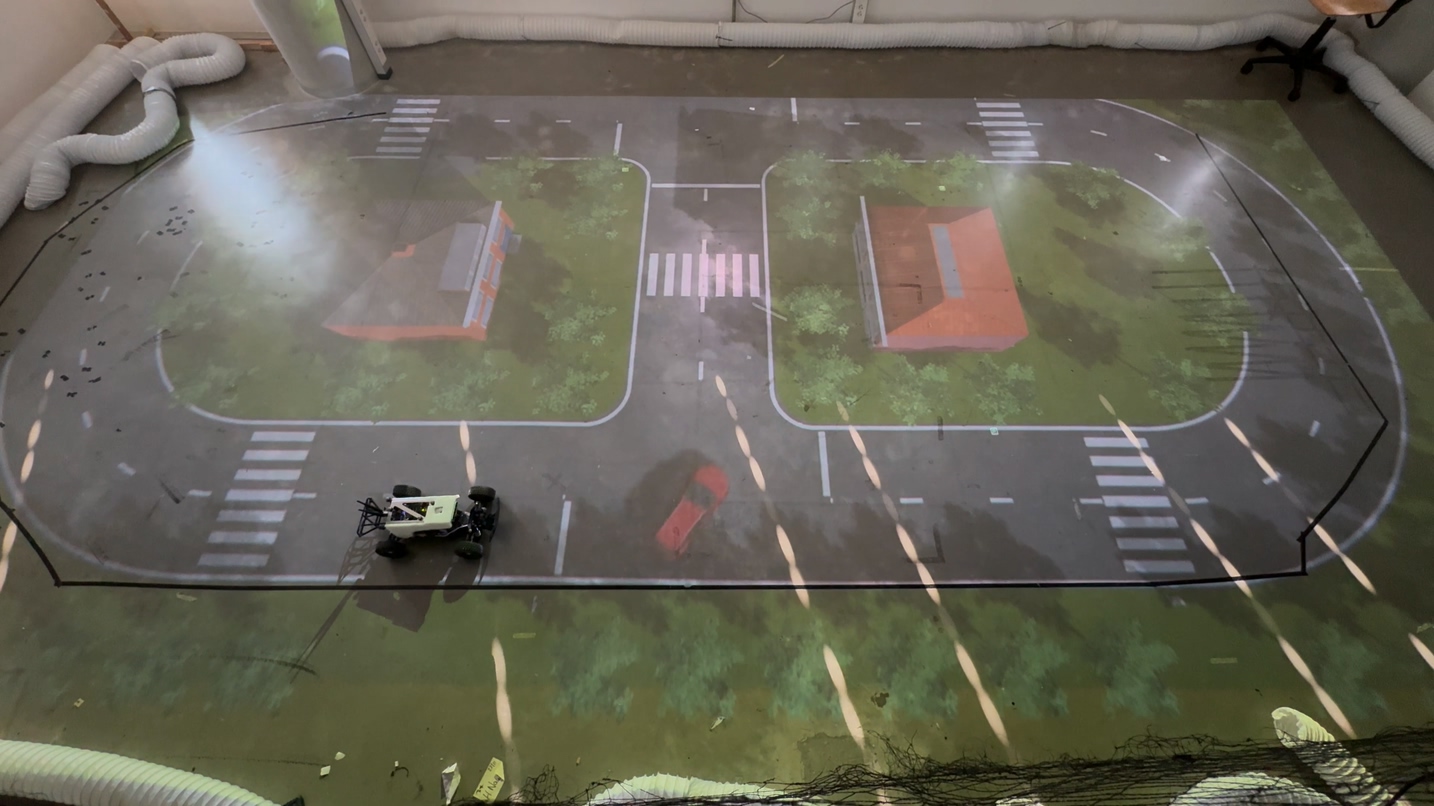}};
\node[minimum height=0.4cm, minimum width=0.4cm, draw, thick, blue] () at (-0.8cm,-0.4cm) {};
\node[minimum height=0.4cm, minimum width=0.4cm, draw, thick, red] () at (-0.1cm,-0.3cm) {};
\end{tikzpicture}
\label{fig:2a}
}
\hspace{-0.4cm}
\subfloat[]{
\begin{tikzpicture}[scale=1,font=\small]%
\node[] (host) at (0cm,0cm) {\includegraphics[height=2.3cm]{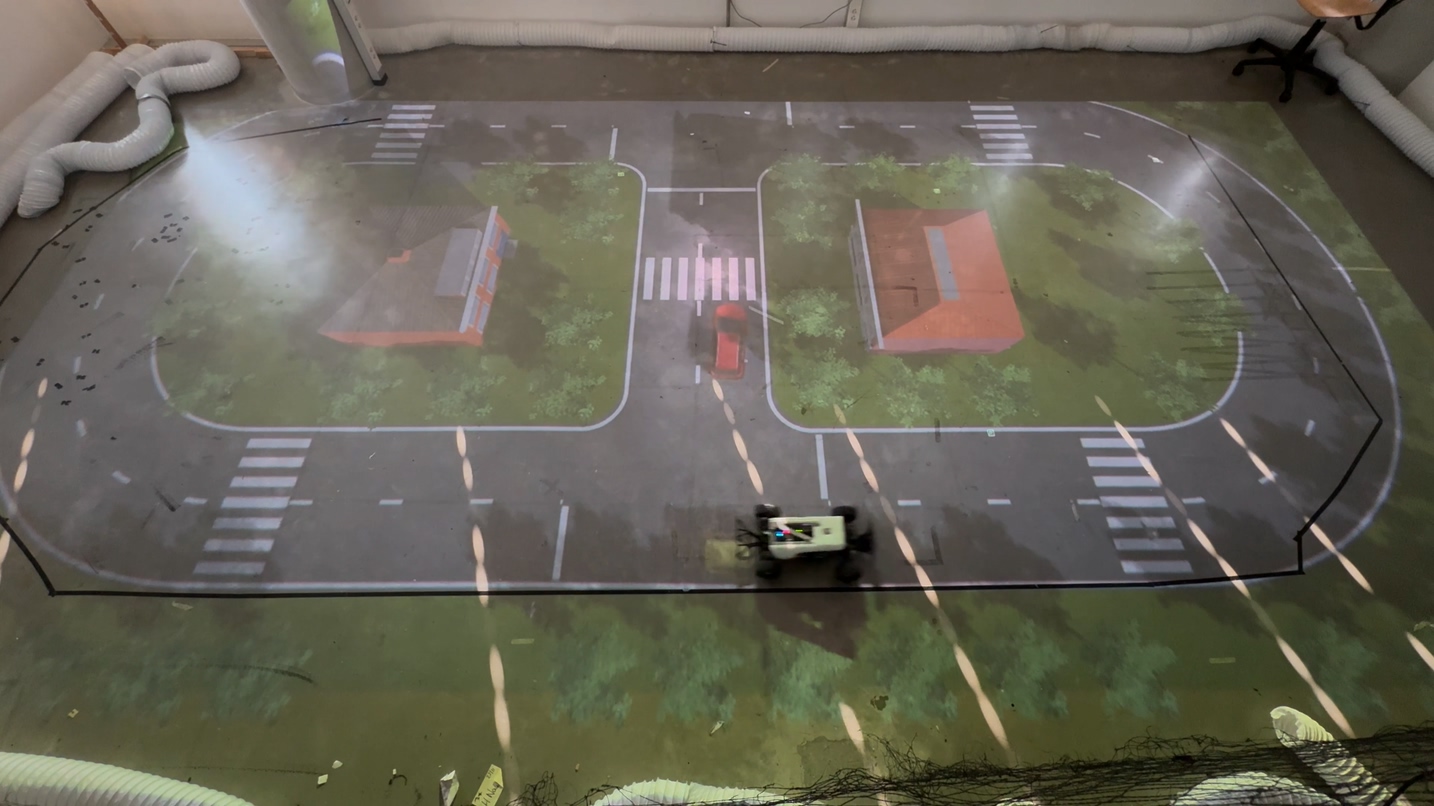}};
\node[minimum height=0.4cm, minimum width=0.4cm, draw, thick, blue] () at (0.2cm,-0.5cm) {};
\node[minimum height=0.4cm, minimum width=0.4cm, draw, thick, red] () at (0cm,0.2cm) {};
\end{tikzpicture}
\label{fig:2b}
}
\hspace{-0.4cm}
\subfloat[]{
\begin{tikzpicture}[scale=1,font=\small]%
\node[] (host) at (0cm,0cm) {\includegraphics[height=2.3cm]{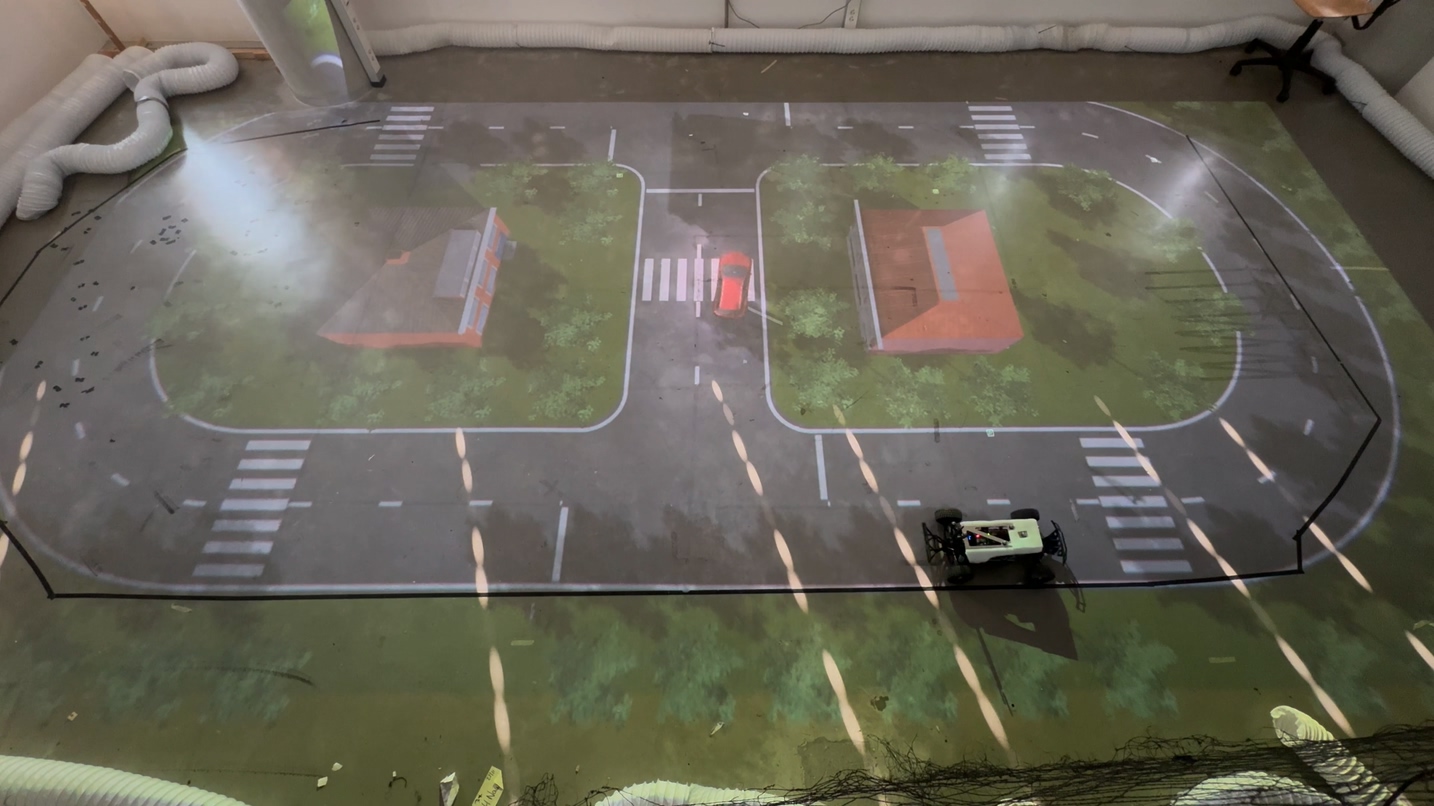}};
\node[minimum height=0.4cm, minimum width=0.4cm, draw, thick, blue] () at (0.8cm,-0.4cm) {};
\node[minimum height=0.4cm, minimum width=0.4cm, draw, thick, red] () at (0cm,0.3cm) {};
\end{tikzpicture}
\label{fig:2c}
}
\hspace{-0.4cm}
\subfloat[]{
\begin{tikzpicture}[scale=1,font=\small]%
\node[] (host) at (0cm,0cm) {\includegraphics[height=2.3cm]{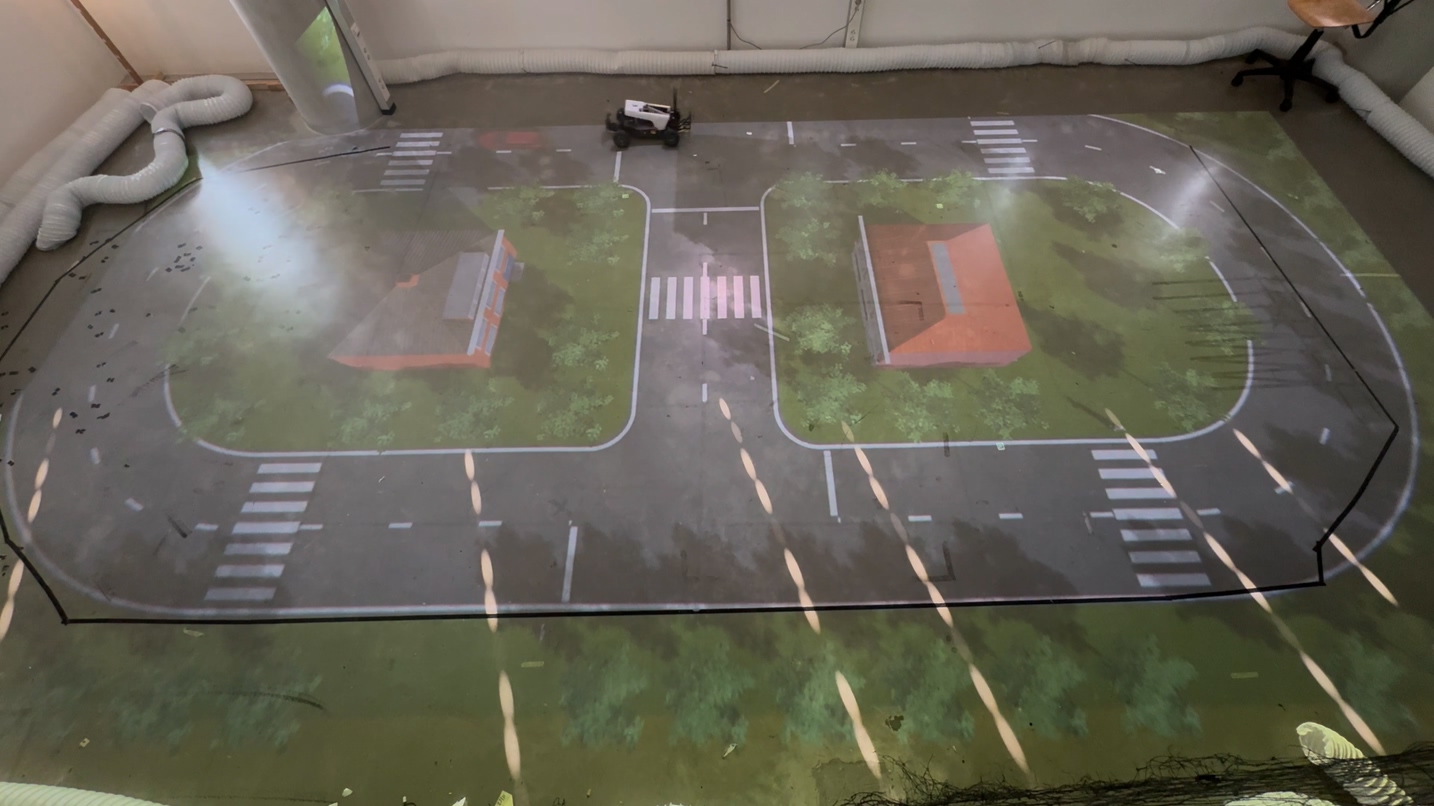}};
\node[minimum height=0.4cm, minimum width=0.4cm, draw, thick, blue] () at (-0.15cm,0.8cm) {};
\node[minimum height=0.4cm, minimum width=0.4cm, draw, thick, red] () at (-0.6cm,0.8cm) {};
\end{tikzpicture}
\label{fig:2d}
}
\caption{Experiment to evaluate the simulator and the safe controller in a scenario with interactive vehicles. (a) The real car (embraced by a blue frame) stops when confronted with a virtual car (embraced by a red frame) to avoid collisions. (b) The real car resumes moving when the virtual car drives away. (c) The real car stops for a pedestrian crossing the road. (d) The real car brakes when seeing the virtual car driving in front, and maintains a certain distance.}
\label{fig:views2}
\end{figure}
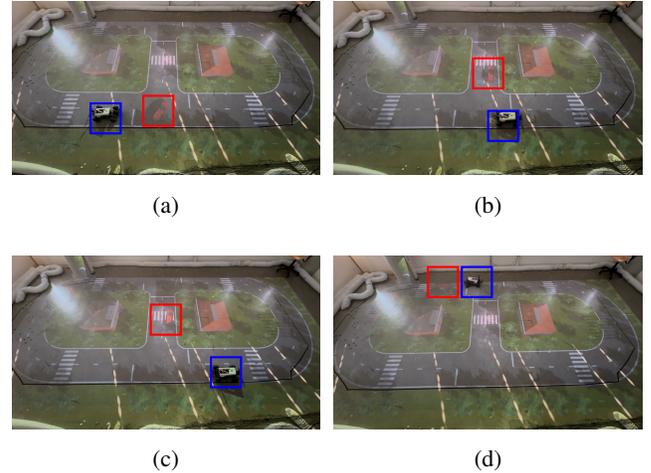

The experiment in Sec.~\ref{sec:val_dt} has verified the precision of the RNN-based DT model. The results also show that the safety filter ensures that the vehicles comply with traffic rules and can follow the predefined paths. In this subsection, we design an experiment with interactive vehicles to showcase the efficacy of the simulator for more complex scenarios with interactive vehicles. This experiment uses the same map as Fig.~\ref{fig:map}, with a scaled F1tenth car as the ego vehicle and a RNN-based virtual model as  an opponent vehicle, as shown in Fig.~\ref{fig:views2}. Different from the first experiment, both vehicles drive along the same track, making it necessary for the ego vehicle to avoid collisions with the opponent vehicle. This is achieved by the safe controller equipped to the ego vehicle.

Fig.~\ref{fig:views2} shows that the ego vehicle is able to follow the desired track and avoid collisions with the opponent vehicle moving in front of it, while following traffic rules. Fig.~\ref{fig:2a} shows that the ego vehicle stops when the front vehicle  turns to the left branch of the track. Fig.~\ref{fig:2b} shows that the ego vehicle resume moving when the collision threats are eliminated. Fig.~\ref{fig:2c} implies that the ego vehicle is able to yield to the pedestrian who is crossing the road. Fig.~\ref{fig:2d} shows that the ego vehicle decelerates when seeing a slowly moving vehicle in front, and keeps a safe distance with it. Thus, this experiment has shown the efficacy of the designed safe controller using a simple but comprehensive scenario containing T-shape intersections, interactive vehicles, and pedestrians. This indicates that the developed ViL simulator is sufficient to evaluate the performance of autonomous vehicles with essential traffic settings, although the complexity of the traffic scenario is limited by the size of the lab.

\section{Conclusion}\label{sec:con}

This paper presents a simulator for validating automated driving control systems, providing essential functions of ViL tests and high-fidelity virtual vehicle models. 
The developed ViL simulator proves effective in simulating essential traffic scenarios. The main advantage of the simulator is its flexibility and versatility, reflected by the possible combination of real and virtual interactive vehicles. The scaled F1tenth car allows ViL tests in limited space. The virtual model based on deep learning technology can be used to perform high-fidelity simulation without hardware. As a result, the simulator also shows its strength in scalability, making it applicable to large-scale traffic scenarios without the limitation of the number of hardware vehicles. Our approach used to developing precise virtual vehicle models can also be extended to the building of simulation models for autonomous systems, such as robots or intelligent industrial processes.

The main limitation of our work is that the complexity of the traffic scenario is still restricted by the lab space. Given a larger lab space, the simulator can be extended by including more infrastructure and traffic participants. Otherwise, a possible solution can be reusing the lab space by constructing folded maps. Another limitation of the current simulator is that the vision functionality is not incorporated, making it difficult to test vision-based control systems. Our future work will focus on developing vision and perception modules for the simulator, including high-fidelity sensors and vision-based control benchmarks. 


\bibliographystyle{IEEEtran}
\bibliography{IEEEabrv, reference.bib}

\begin{thebibliography}{10}
\providecommand{\url}[1]{#1}
\csname url@samestyle\endcsname
\providecommand{\newblock}{\relax}
\providecommand{\bibinfo}[2]{#2}
\providecommand{\BIBentrySTDinterwordspacing}{\spaceskip=0pt\relax}
\providecommand{\BIBentryALTinterwordstretchfactor}{4}
\providecommand{\BIBentryALTinterwordspacing}{\spaceskip=\fontdimen2\font plus
\BIBentryALTinterwordstretchfactor\fontdimen3\font minus \fontdimen4\font\relax}
\providecommand{\BIBforeignlanguage}[2]{{%
\expandafter\ifx\csname l@#1\endcsname\relax
\typeout{** WARNING: IEEEtran.bst: No hyphenation pattern has been}%
\typeout{** loaded for the language `#1'. Using the pattern for}%
\typeout{** the default language instead.}%
\else
\language=\csname l@#1\endcsname
\fi
#2}}
\providecommand{\BIBdecl}{\relax}
\BIBdecl

\bibitem{yurtsever2020survey}
E.~Yurtsever, J.~Lambert, A.~Carballo, and K.~Takeda, ``A survey of autonomous driving: Common practices and emerging technologies,'' \emph{IEEE access}, vol.~8, pp. 58\,443--58\,469, 2020.

\bibitem{wu2021deep}
Y.~Wu, S.~Liao, X.~Liu, Z.~Li, and R.~Lu, ``Deep reinforcement learning on autonomous driving policy with auxiliary critic network,'' \emph{IEEE transactions on neural networks and learning systems}, vol.~34, no.~7, pp. 3680--3690, 2021.

\bibitem{lindemann2023risk}
L.~Lindemann, L.~Jiang, N.~Matni, and G.~J. Pappas, ``Risk of stochastic systems for temporal logic specifications,'' \emph{ACM Transactions on Embedded Computing Systems}, vol.~22, no.~3, pp. 1--31, 2023.

\bibitem{cui2023drivellm}
Y.~Cui, S.~Huang, J.~Zhong, Z.~Liu, Y.~Wang, C.~Sun, B.~Li, X.~Wang, and A.~Khajepour, ``Drivellm: Charting the path toward full autonomous driving with large language models,'' \emph{IEEE Transactions on Intelligent Vehicles}, 2023.

\bibitem{ngo2021multi}
A.~Ngo, M.~P. Bauer, and M.~Resch, ``A multi-layered approach for measuring the simulation-to-reality gap of radar perception for autonomous driving,'' in \emph{2021 IEEE International Intelligent Transportation Systems Conference (ITSC)}.\hskip 1em plus 0.5em minus 0.4em\relax IEEE, 2021, pp. 4008--4014.

\bibitem{hu2023simulation}
X.~Hu, S.~Li, T.~Huang, B.~Tang, R.~Huai, and L.~Chen, ``How simulation helps autonomous driving: A survey of sim2real, digital twins, and parallel intelligence,'' \emph{IEEE Transactions on Intelligent Vehicles}, 2023.

\bibitem{zhang2022high}
Z.~Zhang, R.~Dershan, A.~M.~S. Enayati, M.~Yaghoubi, D.~Richert, and H.~Najjaran, ``A high-fidelity simulation platform for industrial manufacturing by incorporating robotic dynamics into an industrial simulation tool,'' \emph{IEEE Robotics and Automation Letters}, vol.~7, no.~4, pp. 9123--9128, 2022.

\bibitem{grigorescu2020survey}
S.~Grigorescu, B.~Trasnea, T.~Cocias, and G.~Macesanu, ``A survey of deep learning techniques for autonomous driving,'' \emph{Journal of field robotics}, vol.~37, no.~3, pp. 362--386, 2020.

\bibitem{kusari2022enhancing}
A.~Kusari, P.~Li, H.~Yang, N.~Punshi, M.~Rasulis, S.~Bogard, and D.~J. LeBlanc, ``Enhancing sumo simulator for simulation based testing and validation of autonomous vehicles,'' in \emph{2022 ieee intelligent vehicles symposium (IV)}.\hskip 1em plus 0.5em minus 0.4em\relax IEEE, 2022, pp. 829--835.

\bibitem{dosovitskiy2017carla}
A.~Dosovitskiy, G.~Ros, F.~Codevilla, A.~Lopez, and V.~Koltun, ``Carla: An open urban driving simulator,'' in \emph{Conference on robot learning}.\hskip 1em plus 0.5em minus 0.4em\relax PMLR, 2017, pp. 1--16.

\bibitem{chen2020stabilization}
S.~Chen, M.~Wang, W.~Song, Y.~Yang, Y.~Li, and M.~Fu, ``Stabilization approaches for reinforcement learning-based end-to-end autonomous driving,'' \emph{IEEE Transactions on Vehicular Technology}, vol.~69, no.~5, pp. 4740--4750, 2020.

\bibitem{ortega2020overtaking}
J.~Ortega, H.~Lengyel, and Z.~Szalay, ``Overtaking maneuver scenario building for autonomous vehicles with prescan software,'' \emph{Transportation Engineering}, vol.~2, p. 100029, 2020.

\bibitem{li2024choose}
Y.~Li, W.~Yuan, S.~Zhang, W.~Yan, Q.~Shen, C.~Wang, and M.~Yang, ``Choose your simulator wisely: A review on open-source simulators for autonomous driving,'' \emph{IEEE Transactions on Intelligent Vehicles}, 2024.

\bibitem{bullock2004hardware}
D.~Bullock, B.~Johnson, R.~B. Wells, M.~Kyte, and Z.~Li, ``Hardware-in-the-loop simulation,'' \emph{Transportation Research Part C: Emerging Technologies}, vol.~12, no.~1, pp. 73--89, 2004.

\bibitem{landolfi2023hardware}
E.~Landolfi, A.~Salvi, A.~Troiano, and C.~Natale, ``Hardware-in-the-loop validation of an adaptive model predictive control on a connected and automated vehicle,'' \emph{International Journal of Adaptive Control and Signal Processing}, vol.~37, no.~6, pp. 1459--1491, 2023.

\bibitem{albers2010implementation}
A.~Albers and T.~D{\"u}ser, ``Implementation of a vehicle-in-the-loop development and validation platform,'' in \emph{FISITA World automotive congress}, vol. 2010, 2010.

\bibitem{chen2020mixed}
Y.~Chen, S.~Chen, T.~Xiao, S.~Zhang, Q.~Hou, and N.~Zheng, ``Mixed test environment-based vehicle-in-the-loop validation-a new testing approach for autonomous vehicles,'' in \emph{2020 IEEE intelligent vehicles symposium (IV)}.\hskip 1em plus 0.5em minus 0.4em\relax IEEE, 2020, pp. 1283--1289.

\bibitem{szalay20205g}
Z.~Szalay, D.~Ficzere, V.~Tihanyi, F.~Magyar, G.~So{\'o}s, and P.~Varga, ``5g-enabled autonomous driving demonstration with a v2x scenario-in-the-loop approach,'' \emph{Sensors}, vol.~20, no.~24, p. 7344, 2020.

\bibitem{zhen2020control}
Y.~Zhen, Z.~Wang, J.~Liu, T.~Guan, Y.~Zhang, M.~Wang, X.~Song, and D.~Zhang, ``Control strategy and simulation verification of hardware-in-the-loop system of automotive steering device,'' in \emph{IOP Conference Series: Materials Science and Engineering}, vol. 892, no.~1.\hskip 1em plus 0.5em minus 0.4em\relax IOP Publishing, 2020, p. 012048.

\bibitem{rathore2021role}
M.~M. Rathore, S.~A. Shah, D.~Shukla, E.~Bentafat, and S.~Bakiras, ``The role of ai, machine learning, and big data in digital twinning: A systematic literature review, challenges, and opportunities,'' \emph{IEEE Access}, vol.~9, pp. 32\,030--32\,052, 2021.

\bibitem{semeraro2021digital}
C.~Semeraro, M.~Lezoche, H.~Panetto, and M.~Dassisti, ``Digital twin paradigm: A systematic literature review,'' \emph{Computers in Industry}, vol. 130, p. 103469, 2021.

\bibitem{wang2022automatic}
S.-H. Wang, C.-H. Tu, and J.-C. Juang, ``Automatic traffic modelling for creating digital twins to facilitate autonomous vehicle development,'' \emph{Connection Science}, vol.~34, no.~1, pp. 1018--1037, 2022.

\bibitem{yun2021virtualization}
H.~Yun and D.~Park, ``Virtualization of self-driving algorithms by interoperating embedded controllers on a game engine for a digital twining autonomous vehicle,'' \emph{Electronics}, vol.~10, no.~17, p. 2102, 2021.

\bibitem{radanliev2022digital}
P.~Radanliev, D.~De~Roure, R.~Nicolescu, M.~Huth, and O.~Santos, ``Digital twins: artificial intelligence and the iot cyber-physical systems in industry 4.0,'' \emph{International Journal of Intelligent Robotics and Applications}, vol.~6, no.~1, pp. 171--185, 2022.

\bibitem{wu2022digital}
J.~Wu, X.~Wang, Y.~Dang, and Z.~Lv, ``Digital twins and artificial intelligence in transportation infrastructure: Classification, application, and future research directions,'' \emph{Computers and Electrical Engineering}, vol. 101, p. 107983, 2022.

\bibitem{veledar2019digital}
O.~Veledar, V.~Damjanovic-Behrendt, and G.~Macher, ``Digital twins for dependability improvement of autonomous driving,'' in \emph{European conference on software process improvement}.\hskip 1em plus 0.5em minus 0.4em\relax Springer, 2019, pp. 415--426.

\bibitem{lv2022artificial}
Z.~Lv and S.~Xie, ``Artificial intelligence in the digital twins: State of the art, challenges, and future research topics,'' \emph{Digital Twin}, vol.~1, p.~12, 2022.

\bibitem{zhang2024Prescan}
G.~Badakis, M.~Galanis, and A.~Baversi, ``Prescan simulation demonstration,'' \url{https://github.com/zhang-zengjie/prescan-sim-demo}, 2024, gitHub repository.

\bibitem{correia2023data}
J.~B. Correia, M.~Abel, and K.~Becker, ``Data management in digital twins: a systematic literature review,'' \emph{Knowledge and Information Systems}, vol.~65, no.~8, pp. 3165--3196, 2023.

\bibitem{akhmedova2024structures}
Z.~Akhmedova, ``Structures of small database management systems,'' \emph{Solution of social problems in management and economy}, vol.~3, no.~1, pp. 97--107, 2024.

\bibitem{wang2020novel}
R.~Wang, Y.~Li, J.~Fan, T.~Wang, and X.~Chen, ``A novel pure pursuit algorithm for autonomous vehicles based on salp swarm algorithm and velocity controller,'' \emph{IEEE Access}, vol.~8, pp. 166\,525--166\,540, 2020.

\bibitem{marsden2001towards}
G.~Marsden, M.~McDonald, and M.~Brackstone, ``Towards an understanding of adaptive cruise control,'' \emph{Transportation Research Part C: Emerging Technologies}, vol.~9, no.~1, pp. 33--51, 2001.

\bibitem{coelingh2010collision}
E.~Coelingh, A.~Eidehall, and M.~Bengtsson, ``Collision warning with full auto brake and pedestrian detection-a practical example of automatic emergency braking,'' in \emph{13th International IEEE Conference on Intelligent Transportation Systems}.\hskip 1em plus 0.5em minus 0.4em\relax IEEE, 2010, pp. 155--160.

\bibitem{dang2023incorporating}
N.~Dang, Z.~Zhang, J.~Liu, M.~Leibold, and M.~Buss, ``Incorporating target vehicle trajectories predicted by deep learning into model predictive controlled vehicles,'' \emph{arXiv preprint arXiv:2310.02843}, 2023.

\bibitem{zhang2024data}
\BIBentryALTinterwordspacing
Z.~Zhang, G.~Badakis, and M.~Galanis, ``Driving data of a real f1tenth car,'' 2024. [Online]. Available: \url{https://doi.org/10.5281/zenodo.12536535}
\BIBentrySTDinterwordspacing

\bibitem{zhang2024modeling}
Z.~Zhang, ``Modeling and motion prediction of autonomous vehicles using deep learning,'' \url{https://github.com/zhang-zengjie/dl-vehicle-model}, 2024, gitHub repository.

\bibitem{calin2017book}
C.~Belta, B.~Yordanov, and E.~Gol, \emph{Formal Methods for Discrete-Time Dynamical Systems}.\hskip 1em plus 0.5em minus 0.4em\relax Springer, 01 2017, vol.~89.

\bibitem{filippidis2016control}
I.~Filippidis, S.~Dathathri, S.~C. Livingston, N.~Ozay, and R.~M. Murray, ``Control design for hybrid systems with tulip: The temporal logic planning toolbox,'' in \emph{2016 IEEE Conference on Control Applications (CCA)}.\hskip 1em plus 0.5em minus 0.4em\relax IEEE, 2016, pp. 1030--1041.

\bibitem{zhang2024sm}
Z.~Zhang, ``State machine generation for autonomous driving,'' \url{https://github.com/zhang-zengjie/sm-for-driving}, 2024, gitHub repository.

\bibitem{maierhofer2022formalization}
S.~Maierhofer, P.~Moosbrugger, and M.~Althoff, ``Formalization of intersection traffic rules in temporal logic,'' in \emph{2022 IEEE Intelligent Vehicles Symposium (IV)}.\hskip 1em plus 0.5em minus 0.4em\relax IEEE, 2022, pp. 1135--1144.

\end{thebibliography}

\end{document}